\documentclass[11pt]{article}

\usepackage[final]{acl}

\usepackage{times}
\usepackage{latexsym}

\usepackage[T1]{fontenc}
\usepackage[utf8]{inputenc}

\usepackage{microtype}

\usepackage{inconsolata}

\usepackage{graphicx}
\usepackage[export]{adjustbox}

\usepackage{natbib}

\usepackage{enumerate}   

\usepackage[T1]{fontenc}
\usepackage[utf8]{inputenc}
\usepackage{graphicx}
\graphicspath{ {./images/} }

\usepackage{verbatim}

\usepackage{amsmath}
\usepackage{array}
\usepackage{booktabs}
\usepackage{multirow}
\usepackage{appendix}

\usepackage{makecell}

\usepackage{float}
\usepackage{tabularx}

\usepackage[gen]{eurosym}

\usepackage{subcaption}
\usepackage{threeparttable}

\usepackage[hang,flushmargin]{footmisc}

\title{Leveraging LLMs for Context-Aware Implicit\\ Textual and Multimodal Hate Speech Detection}

\author{
  \textbf{Joshua Wolfe Brook}, \textbf{Ilia Markov}\\
  Computational Linguistics \& Text Mining Lab,\\
  Vrije Universiteit Amsterdam\\
  \texttt{j.w.brook@student.vu.nl}, \texttt{i.markov@vu.nl}
}

\begin{document}
\maketitle
\begin{abstract}

This paper investigates the use of an LLM to generate auxiliary background context for social media posts, and explores four methods to incorporate this context into the input of an SBERT-based Hate Speech Detection (HSD) classifier. These are: text concatenation, embedding concatenation, a hierarchical transformer-based fusion, and LLM-driven text enhancement. We evaluate the impact of our context generation and incorporation strategies in a textual setting on the \textit{Latent Hatred} dataset of implicitly hateful tweets and a multimodal setting on the \textit{MAMI} dataset of misogynous internet memes. Results are evaluated against a zero-context baseline, two previous approaches based on entity linking, and a zero-shot LLM classifier.
Findings indicate that incorporating generated context improves HSD performance by up to 3 and 6 F1 points on textual and multimodal settings respectively, from a zero-context baseline to the highest-performing system, based on embedding concatenation.

\textbf{Content Warning:} This paper contains examples of hate speech, including explicit language and content that may be racist, sexist, or otherwise offensive.
\end{abstract}

\section{Introduction}

Detecting hate speech in social media posts is a complex and context-dependent task, as such speech is often obfuscated through irony, euphemism, and coded language \cite{hendersonHowDogwhistlesWork2018, bourgeadeHumansNeedContext2024}, or its understanding is reliant on external world knowledge \cite{sood-memhatecaptioning}. While current models perform relatively well at flagging explicit hate speech, implicit hate, where hostility is veiled or implied, remains far more difficult to detect \cite{elsheriefLatentHatred2021, ocampo-etal-2023-depth}.
Hate Speech Detection (HSD) has most frequently been performed in zero-context settings \cite{pavlopoulos-2020-toxicity, gaoDetectingOnlineHate2017}, predicting whether content is hateful based solely on the representation of a raw post. While this approach simplifies model design and aligns with the structure of many available datasets, it can often fail to capture the underlying intent behind implicitly hateful language. 
One method used to improve implicit HSD systems is to incorporate additional context into the model input \cite{markovRoleContextDetecting2022, bourgeadeHumansNeedContext2024}. This context can be identified from the surrounding conversation (\textit{Conversational Context} e.g., replies to a target post) \cite{gaoDetectingOnlineHate2017, markovRoleContextDetecting2022, perezAssessingImpactContextual2023} or through augmenting target text with externally-sourced world knowledge (\textit{Background Context}) \cite{elsheriefLatentHatred2021, linLeveragingWorldKnowledge2022}. The former is not always feasible to incorporate, as available labelled datasets are generally composed of isolated posts without any surrounding discourse \cite{elsheriefLatentHatred2021}, and the latter methodology often struggles to identify relevant information due to poor entity detection or irrelevant linked data \cite{linLeveragingWorldKnowledge2022}.

Addressing the aforementioned limitations, this research investigates how Large Language Models (LLMs) can serve as dynamic knowledge bases to generate background information about online posts that may contain (implicit) hate speech. While prior studies have integrated generative LLMs into the HSD pipeline, these models have typically been utilised as direct replacements for classifiers, leading to mixed results \cite{guoInvestigationLargeLanguage2024, chiu2022detectinghatespeechgpt3, royProbingLLMsHate2023}.
In contrast, we focus on leveraging the rich contextual knowledge embedded in pre-trained LLMs, without abandoning the simplicity and robustness of traditional classification models. 
As such, rather than relying on generic pre-structured external resources (e.g., ConceptNet \cite{speer2017conceptnet}), we explore prompting an LLM, \textit{Gemini 2.0 Flash} \cite{gemini2flash} to generate auxiliary background knowledge specific to each post, based on its entire context, rather than solely its extracted entities, as in previous studies \cite{elsheriefLatentHatred2021, linLeveragingWorldKnowledge2022}.

Further, we test several methods to incorporate this context into the input of a Multi-Layer Perceptron (MLP) classifier and evaluate performance across various experimental setups.
Approaches are first tested on the \textit{Latent Hatred} \cite{elsheriefLatentHatred2021} dataset of implicitly hateful Twitter posts in a binary hate or non-hate classification setup, followed by multi-class identification of fine-grained implicit hate speech categories (e.g., \textit{incitement}, \textit{stereotyping}), based on \citeposs{elsheriefLatentHatred2021} typology for ``characterising and detecting different forms of implicit hate''. Additionally, we adapt our approaches to a multimodal setting, using \citeposs{fersiniMAMI2022} \textit{Multimedia Automatic Misogyny Identification} (MAMI) dataset of misogynous memes, to determine how well our approach generalises to a new setting and domain.

Overall, we investigate whether the shortcomings of previous context-aware HSD systems are due to the quality of the context itself or the method by which it is incorporated, and address the following research questions: 
(1) To what extent is an LLM more effective than previous static entity linking approaches at generating background context for implicit and multimodal HSD? 
(2) Which method of incorporating LLM-generated background context yields the greatest performance in a downstream HSD model?
(3) What are the benefits and trade-offs of using LLM-generated background context compared to direct LLM classification across textual and multimodal HSD?

\section{Related Work}

While hate speech is inherently contextual \cite{markovRoleContextDetecting2022, pavlopoulos-2020-toxicity}, previous studies have shown that incorporating generic additional context into the model input often leads to a negligible or negative impact on model performance \cite{meniniAbuseContextualWhat2021, perezAssessingImpactContextual2023}. These findings suggest that context must be carefully selected, represented, and integrated to avoid introducing noise. This section discusses previous work relevant to HSD in context retrieval and incorporation, the use of LLMs, and the challenges surrounding multimodality.

\subsection{Conversational Context}

Conversational context refers to content structurally or temporally linked to the target post, such as parent posts or prior replies. \citet{gaoDetectingOnlineHate2017} incorporated usernames and article titles into their dataset of hateful comments left in response to said articles to improve F1 scores of a binary HSD model by 5 points. \citet{markovRoleContextDetecting2022} added previous replies to Facebook comments targeting migrants, achieving a 6-point F1 gain in detecting the targets of hate speech. \citet{perezAssessingImpactContextual2023} used news article titles and prior comments, boosting binary classification F1 by 2 points and multi-class HSD by 6. Such context can provide thematic grounding and discourse cues, though its relevance can vary and its incorporation must be carefully considered \cite{meniniAbuseContextualWhat2021}.

\subsection{Background Context}

In addition to, or when conversational context is unavailable, background context can be added to augment the post with world knowledge and make the post easier to understand \cite{sharifiradBoostingTextClassification2018}. Typically, entity linking has been used to retrieve background knowledge related to named entities from external sources \cite{elsheriefLatentHatred2021, linLeveragingWorldKnowledge2022}.
\citet{sharifiradBoostingTextClassification2018} augment sexist Tweets with knowledge extracted from ConceptNet \cite{speer2017conceptnet} and \citet{wikidata}, boosting prediction accuracy by ca. 5 points. \citet{elsheriefLatentHatred2021} test a similar approach on their own dataset, also extracting entities and linking to ConceptNet and Wikidata to incorporate background context about a wide range of entities. 
\citet{linLeveragingWorldKnowledge2022} expands on this, using the Radboud Entity Linker (REL) \cite{vanHulstREL2020} to extract Wikipedia descriptions of named entities, concatenating these with tweets and encoding them via SBERT \cite{reimers-2019-sentence-bert}. This approach improved binary HSD macro F1 by 10 points but reduced multi-class performance on fine-grained hate categories by 11 points.\footnote{Given that this is one of the baseline approaches used in our study, we discuss some potential shortcomings of REL and Lin's methodology in Appendix \ref{rel-note}.}

\subsection{Large Language Models}

LLMs have been used in HSD previously, generally replacing classifiers themselves, as in \citet{guoInvestigationLargeLanguage2024}, who directly compare a generative LLM against two pre-trained models: BERT \cite{bert-devlin-2019} and RoBERTa \cite{roberta}. They test various prompting strategies (few-shot, chain-of-thought, etc.) and demonstrate that a generalist LLM with simple prompts can comfortably outperform transformer-based encoders. These results are not confirmed, however, by numerous other studies, who find that LLMs either match or fall short of the performance of encoder-only classifiers, especially without extensive prompt engineering \cite{chiu2022detectinghatespeechgpt3, royProbingLLMsHate2023}. \citet{albladi-llm-hsd} discuss these results, highlighting the fact that encoder models are fine-tuned on the training data and can potentially capture the peculiarities of a given dataset better than generalist LLMs. 
% This is especially relevant when working in a multilingual setting, as demonstrated by \citet{dehghan-yanikoglu-2024-evaluating} and \citet{ghorbanpour2025promptingllmsunlockhate}.

While replacing traditional classifiers entirely is a promising approach, this research instead focuses on treating LLMs as dynamic knowledge bases from which to extract contextual information. \citet{petroniLLMsAsKBs2019} were among the first to systematically evaluate whether generative models can store factual and relational knowledge. Using cloze-style prompts, they demonstrate that pre-trained models can recall factual information with high accuracy, often outperforming traditional relation extraction systems.
Building on this idea, \citet{wangKGsFromLLMs2020} propose a method for building structured knowledge graphs directly from LLMs, suggesting that larger models can be used to effectively store and retrieve world knowledge. These successes motivate our approach of using LLMs to retrieve post-specific background context to enhance HSD performance.

\subsection{Incorporating Context}

Previous approaches often integrate context by simply appending it to the target text \cite{gaoDetectingOnlineHate2017, markovRoleContextDetecting2022, perezAssessingImpactContextual2023}, relying on the embedding model to distinguish content from context. \citet{meniniAbuseContextualWhat2021} take such a route, re-annotating a dataset from \citet{fountaLargeScaleCrowdsourcing2018} in a context-aware setting. Providing conversational context to annotators reduced the number of posts labelled as abusive (from 18\% to 10\% of the data). Using this re-annotated dataset, the authors append all posts deemed contextually relevant to each target post before classification. This approach reduced performance, demonstrating that naively adding large amounts of context can do more harm than good to a HSD system. Similar results were observed by \citet{pavlopoulos-2020-toxicity}, who investigate whether context is as important as is often assumed in HSD.
\citet{bourgeadeHumansNeedContext2024} propose a novel approach to incorporate conversational context into representations of Tweets. This approach, dubbed \textit{Context-Embed}, passes the context through an encoder-only model and the post through the word embedding matrix of a separate transformer, before concatenating the two and passing the unified representation through the rest of the second transformer's layers. This method was shown to generally outperform classification based on text concatenation, leading us to implement a variation of it for one of our own experimental setups.

\subsection{Multimodality}

Detecting hate speech in multimodal formats (e.g., memes, which combine visual and textual elements) shares many of the same challenges as implicit HSD \cite{pandianiToxicMemesSurvey2024}. In both cases, the hateful intent can be subtle, obfuscated, or reliant on external context, making it difficult to identify using surface-level cues alone \cite{sood-memhatecaptioning}. Often, a meme's text may appear benign until paired with a provocative image, or vice versa \cite{kiela2021hatefulmemeschallengedetecting}, just as an implicitly hateful post may seem innocuous without background knowledge or cultural awareness. This conceptual overlap makes multimodal HSD a relatively natural extension of implicit HSD and a compelling setting for evaluating the effectiveness of context-enhanced models.

\section{Methodology}

This research explores the potential of LLMs to serve as dynamic knowledge bases for retrieving and integrating relevant background context into (hateful) online posts. The methodology involves two core stages: context generation and context incorporation, tested in both textual and multimodal setups. To generate context, we provide the full text of each post to an LLM with a simple prompt.\footnote{Full prompt texts are provided in Appendix \ref{tab:prompts}.} To incorporate context, four distinct techniques are tested (described in detail in §\ref{exp}). In the multimodal setting, the full-text pipeline is adapted for memes by extracting image text and generating captions and background context all using an LLM. Sentence embeddings are created using SBERT, and are passed to an MLP classifier. 
Overall, this modular design ensures the isolation of the effects of different context generation and integration strategies to assess their impact independently of the classification and representation architectures. We have made our code and context-augmented data publicly available.\footnote{\url{https://github.com/joshbrook/contextual-hsd}}

\subsection{Datasets}

This research employs two open-source, English-language datasets used to train and evaluate our approaches over various setups.

\subsubsection{Latent Hatred}

The first is \citeposs{elsheriefLatentHatred2021} textual \textit{Latent Hatred} dataset, focused on identifying \textit{implicit} hate speech. The authors establish a theoretical framework to identify the ``diverse manifestations'' of implicit hatred, including a six-way multi-class taxonomical classification of implicit hate speech varieties. These include: incitement to violence, inferiority language, irony, stereotyping, threatening, and white grievance. 
This dataset consists of 21,480 tweets from known hate group accounts and is comprised of roughly 62\% non-hateful posts, with 33\% labelled as implicit hate speech and 5\% explicit. The implicit multi-class labels are distributed as in Table \ref{tab:class_dist} of Appendix \ref{dists}. We randomly split the data into training and test sets in an 80/20 ratio while maintaining class balance.

\subsubsection{MAMI}

Our second source is \citeposs{fersiniMAMI2022} \textit{Multimedia Automatic Misogyny Identification} (MAMI) dataset of misogynous memes. This dataset is comprised of 10,995 distinct memes\footnote{Five duplicate memes were discovered in MAMI during preprocessing and were subsequently removed.} scraped from social media platforms and meme-creation websites. 
The high-level annotation is a binary distinction between misogynous\footnote{A meme is defined as misogynous if it ``conceptually describes an offensive, sexist or hateful scene [...] having as target a woman or a group of women''.} and non-misogynous memes, with a 50/50 distribution in the dataset.
In the second level annotation, misogynous memes are labelled with one or more subtypes of misogyny: shaming, stereotyping, objectification, and violence. These labels are distributed as in Table \ref{label_dist} of Appendix \ref{dists}, with example memes and their labels given in Appendix \ref{examples}.
We again split the data in an 80/20 ratio for training and evaluation.

\subsection{Model Selection}

For each post, background context is generated, then four combined representations are created, before being classified.

\subsubsection{Large Language Model}

We employ Google's \texttt{Gemini 2.0 Flash} \cite{gemini2flash} model for all experiments; a decision driven primarily by speed and cost-efficiency.
This model can be prompted through a batch API, allowing for processing of many requests in parallel. This was considered essential, given the need to generate context for thousands of instances, often multiple times to experiment with consistency.
While various LLMs offer similar capabilities, Gemini's high performance-to-cost ratio, highlighted by LMArena’s \cite{lmarena} benchmark evaluations, made it an attractive choice.

\subsubsection{Sentence Embeddings}

Posts are encoded using the SBERT  \texttt{all-mpnet-base-v2} model, which creates normalised 768-length embeddings from sentences up to 384 tokens and is the highest-performing pre-trained generalist model from Sentence Transformers \cite{reimers-2019-sentence-bert}.

\subsubsection{Classification Model}

We implement a simple MLP classifier, mirroring the architecture outlined by \citet{linLeveragingWorldKnowledge2022}, used as one of our baselines. The network uses three hidden layers of dimension 512 with ReLU activations, and was trained in each experiment for 500 epochs with the \texttt{Adam} optimiser and a learning rate of 0.001.

\subsection{Baselines} \label{baselines}

To evaluate the effectiveness of our LLM-based context-enhanced approaches and compare against the results obtained by replicating the approaches of previous studies, we establish four baselines.

\begin{enumerate}[i)]
    \item \textbf{Zero-Context:} The first configuration uses only the raw post, simply creating SBERT embeddings from each post in the dataset and feeding them directly into the input of the MLP classification model.
    
    \item \textbf{REL:} The second baseline replicates \citeposs{linLeveragingWorldKnowledge2022} entity-linking strategy, augmenting post text with Wikipedia-derived summaries based on named entities. Posts are processed using REL \cite{vanHulstREL2020}, with identified entities mapped to corresponding Wikipedia articles. The first two sentences of these articles are then appended to the post and their combination is encoded using SBERT.

    \item \textbf{ConceptNet:} The third baseline augments posts using ConceptNet's Numberbatch embeddings \cite{speer2017conceptnet}, replicating the methodology outlined by \citet{elsheriefLatentHatred2021}. Entities are extracted using n-gram matching before being mapped to ConceptNet embeddings. These vectors are then averaged and normalised, forming a single contextual representation which is concatenated with the SBERT embedding of the original text.

    \item \textbf{LLM Classifier:} The final baseline system forgoes embeddings and training entirely, instead using the generative LLM directly as a classifier. 
    % \citet{willats2025classification-rag} use a similar approach to HSD in their research, demonstrating strong performance and high adaptability to changes in policy.
    This setup aims to determine whether, if given access to an LLM, such a model is better utilised for generating context or directly for prediction. The LLM is prompted solely with each post and the potential labels and asked to predict binary, multi-class and multi-label classes.\footnote{Full prompt texts are provided in Appendix \ref{tab:prompts}.}
\end{enumerate}

\subsection{Experiments} \label{exp}

Our experiments are designed to systematically evaluate how post-specific LLM-generated background context impacts HSD performance across textual and multimodal scenarios.

\subsubsection{Context Generation}

As discussed previously, contextualised HSD approaches commonly retrieve external knowledge based on identified named entities. Rather than treating contextualisation as an entity-level task, we instead generate background context based on the full content of each post, allowing the LLM to dynamically determine which information is relevant. This may better resolve ambiguous references and capture implicit meanings that emerge only from the post as a whole.
To reduce the likelihood of implicit label leakage, the prompt is framed as a neutral knowledge-generation task and is not linked to hate speech. The LLM is not informed that its output will be used for classification and is instead asked to produce objective background context, including information on relevant entities, events, and implicit meanings or rhetorical devices.

\subsubsection{Context Incorporation}

Once context has been generated, we test four strategies to incorporate it into the model input, dubbed i) \textit{Append \& Embed}, ii) \textit{Embed \& Concat}, iii) \textit{Context-Embed}, and iv) \textit{LLM Enhance}.

\begin{enumerate}[i)]
    \item \textbf{Append \& Embed} is the approach to integrating context most frequently used in previous studies: appending context directly to the end of the post, following a \texttt{[SEP]} token, before creating embeddings from their union. We expect this may bias the embeddings to represent the context more than the post itself, as generated context is generally much longer than the original text (with an average of 17 tokens per post and 78 per contextual summary).
    
    \item \textbf{Embed \& Concat} tests the inverse of the previous strategy, vectorising the post and context separately, then concatenating them. This preserves a greater separation between post and context, potentially aiding the classifier in distinguishing their respective contributions.

    \item \textbf{Context-Embed} follows \citeposs{bourgeadeHumansNeedContext2024} methodology, fusing post and context hierarchically at the embedding layer before encoding, leading to a deeper integration of context and post than simple concatenation at the input or embedding level. In this configuration, the context is encoded with SBERT, while the post is tokenized and passed through the word embedding matrix of another SBERT model. The context vector is projected to match the dimensionality of token embeddings and prepended to the post embedding sequence. This expanded sequence is then passed through the remaining layers of the encoder. The resulting representation is then pooled before being passed to the classifier.

    \item \textbf{LLM Enhance} is a novel approach to context incorporation, inspired by a study on generation of counter-narratives against hate speech by \citet{doganc-markov-2023-generic}. Here, an LLM is provided with each post and its generated context and prompted to combine them, creating a modified context-enhanced post. This tests whether rephrasing the post to include more context yields better performance than preserving the original structure and adding context separately.

\end{enumerate}

\begin{table*}[ht]
\centering
\begin{threeparttable}
\footnotesize
\centering
\begin{tabular}{p{2.5cm} p{2.5cm}
    >{\centering\arraybackslash}p{1.8cm}
    >{\centering\arraybackslash}p{1.8cm}
    >{\centering\arraybackslash}p{2.4cm}}
    \toprule
    \multirow{2}{*}{\textbf{Context}}
    & \multirow{2}{*}{\makecell[l]{\textbf{Incorporation}\\ \textbf{Strategy}}}
    & \multicolumn{2}{c}{\textbf{Binary}} & \textbf{Multi-Class} \\
    & & \textbf{Macro F1} & \textbf{Hate F1} & \textbf{Macro F1} \\
    \midrule

    \multicolumn{5}{c}{\textbf{Baselines}} \vspace{3pt}\\
    Zero-Context & - & 0.72 & 0.65 & \textbf{0.51} \\
    REL & Append \& Embed & 0.70 & 0.63 & 0.47 \\
    ConceptNet & Embed \& Concat & \textbf{0.74} & \textbf{0.67} & 0.50 \\
    LLM Classifier & - & 0.70 & 0.65 & 0.26 \\

    \midrule
    \multicolumn{5}{c}{\textbf{Experiments}} \vspace{3pt}\\
    \multirow{4}{*}{LLM-Generated}
    & Append \& Embed & 0.73 & 0.66 & \textbf{0.53} \\
    & Embed \& Concat & \textbf{0.75} & \textbf{0.69} & \textbf{0.53} \\
    & Context-Embed & 0.71 & 0.64 & 0.47 \\
    & LLM Enhance & 0.71 & 0.63 & 0.49 \\

    \bottomrule
\end{tabular}
\caption{Macro and positive class (hate) F1 scores for binary HSD and macro F1 scores for implicit multi-class prediction on the Latent Hatred dataset.}
\label{tab:lh-res-main}
\end{threeparttable}
\end{table*}

\subsubsection{Multimodal Approach}

In the multimodal setting, we adapt our context generation and incorporation pipeline to classify memes from the MAMI dataset, in both binary and multi-label setups.
We represent each meme in a textualised way, through the combination of two texts, comprised of (1) the text extracted directly from the meme and (2) a description of the image itself. This is a relatively common approach to representing multimodal input, used previously for HSD by e.g., \citet{das2020detectinghatespeechmultimodal} and \citet{britez2025cltl}. 
Additional details on our choice of representation are provided in Appendix \ref{representation}.
Context extraction mirrors the methodology used in the textual setting, though multimodally, prompting the LLM with each meme directly to generate background context.\footnote{Examples of LLM-generated context are provided in Appendix \ref{examples}, Table \ref{multimodal-examples}.} Context is incorporated using the same strategies described in §\ref{exp}.

\subsection{Evaluation}

Each experiment is run ten times with mean results reported. The primary evaluation metric used in this study is macro F1. We also report positive class F1 in the binary settings and per-class F1 for the multi-class and multi-label settings.

\section{Results}

This section presents the results of our experiments across two datasets: Latent Hatred for textual HSD and MAMI for multimodal misogyny detection. 
%We evaluate the impact of LLM-generated context under multiple integration strategies and compare these against established baselines.

\subsection{Latent Hatred: Textual HSD}

Table \ref{tab:lh-res-main} summarises macro and positive-class F1 scores for binary HSD and multi-class implicit hate speech classification on the Latent Hatred dataset. Our best-performing configuration was the \textit{Embed \& Concat} model, which outperformed all baselines, achieving macro F1 scores of 0.75 in the binary setting and 0.53 in the multi-class setting.\footnote{Full results tables with precision, recall, and per-class scores are provided in Appendix \ref{full-res}.}
Our implementations of the REL and ConceptNet baselines, described in §\ref{baselines}, underperformed the zero-context model in the multi-class setting, though the latter performed better in binary classification.
The REL approach, in particular, lead to a 4-point drop in F1, reaffirming the limitations of static entity-linking approaches given noisy data.
Overall, both \textit{Context-Embed} and \textit{LLM Enhance} performed worse than the baselines, demonstrating that a deeper incorporation of context (i.e., at the text level or embedding level) is less effective than just keeping them separate.
Predictions made directly by the LLM classifier were comparable to other baselines on the binary task, but fell well behind on multi-class prediction. As such, there appears to be utility in using an LLM in the HSD pipeline to generate context, rather than solely as a replacement classifier.

\begin{table*}[h]
\centering
\begin{threeparttable}
    \footnotesize
    \centering
    \begin{tabular}{p{2.5cm} p{2.5cm}
        >{\centering\arraybackslash}p{1.8cm}
        >{\centering\arraybackslash}p{1.8cm}
        >{\centering\arraybackslash}p{2.4cm}}
        \toprule
        \multirow{2}{*}{\textbf{Context}}
        & \multirow{2}{*}{\makecell[l]{\textbf{Incorporation}\\ \textbf{Strategy}}} 
        & \multicolumn{2}{c}{\textbf{Binary}} 
        & \textbf{Multi-Label} \\
        & & \textbf{Macro F1} & \textbf{Hate F1} & \textbf{Macro F1} \\
        \toprule
        
        \multicolumn{5}{c}{\textbf{Baselines}} \vspace{3pt}\\
        Zero-Context & - & 0.79 & 0.79 & 0.59 \\
        REL & Append \& Embed & 0.78 & 0.78 & 0.57 \\
        ConceptNet & Embed \& Concat & 0.80 & 0.80 & \textbf{0.60} \\
        LLM Classifier & - & \textbf{0.86} & \textbf{0.87} & \textbf{0.60} \\
        \midrule

        \multicolumn{5}{c}{\textbf{Experiments}} \vspace{3pt}\\
        \multirow{4}{*}{LLM-Generated} & Append \& Embed & 0.84 & 0.84 & 0.61 \\
        & Embed \& Concat & \textbf{0.85} & \textbf{0.85} & \textbf{0.63} \\
        & Context-Embed & 0.83 & 0.84 & 0.61 \\
        & LLM Enhance & 0.83 & 0.83 & 0.62 \\
        \bottomrule
    \end{tabular}
    \caption{Macro \& positive class (misogynous) F1 scores for binary HSD and macro F1 scores for multi-label classification on the MAMI dataset.}
    \label{mami-res}
    \end{threeparttable}
\end{table*}

\subsection{MAMI: Multimodal HSD}

Table \ref{mami-res} reports macro and positive-class F1 scores for binary HSD and multi-label 
misogyny classification on the MAMI dataset. Results indicate that incorporating LLM-generated context by any method improved performance against all embedding-based baselines in the multimodal setting. 
The \textit{Embed \& Concat} approach again outperformed other strategies, achieving gains of 6 and 4 F1 points over the zero-context baseline in binary and multi-label settings, respectively.
In the multi-label task, we observe slight improvements across the board for all labels, demonstrating that our strategy of incorporating context from an LLM does not majorly impact the distribution of correct predictions, only bringing the minority classes closer to level.\footnote{See Table \ref{mami-extra-results} of Appendix \ref{full-res} for per-label results.}
% This demonstrates that LLM-generated context is particularly valuable in multimodal scenarios.
Once again, the REL model underperformed the zero-context baseline, likely due to limited named entity coverage in memes. We briefly discuss the limitations of multimodal NER in Appendix \ref{no-nes}. 
The ConceptNet approach fared better but remained within 1 F1 point of the baseline in both tasks.
Lastly, the LLM Classifier system performs substantially better in this setting than on Latent Hatred, surpassing the results of all other setups on binary classification, while maintaining lower but still competitive multi-label performance.

%\subsection{Comparative Observations}

%Across both datasets, the clearest pattern is that separating post and context at the embedding level (\textit{Embed \& Concat}) consistently outperforms deeper fusion methods, particularly in binary classification. While simpler fusion strategies (e.g., \textit{Append \& Embed}) showed modest gains, overly complex approaches (e.g., \textit{Context-Embed} in low-context settings) risk performance degradation. Direct LLM prediction performs reasonably in binary tasks but deteriorates sharply in nuanced, multi-class settings, suggesting that generative models alone are still insufficient for fine-grained distinctions without explicit training or advanced prompt engineering. Together, these results indicate that LLM-generated context is most effective when (i) it is drawn from the full input rather than from extracted entities and (ii) it is kept modular and separately weighted rather than fully merged into the text or embeddings.

\section{Error Analysis}

% Error analysis is essential for a task as nuanced as HSD, where meaning is often implied, contextual, and subject to annotation variability. 
To assess the influence of LLM-generated context on model behaviour, we compare variance between the predictions of the zero context model and the best-performing \textit{Embed \& Concat} setup, rather than solely against the true labels. 

%Frequent error patterns in HSD have been observed by, for instance, \citet{van-aken-etal-2018-challenges}, who assert that their ensemble model struggles to identify hate embedded in rhetorical questions, metaphors and irony, while frequently mislabelling non-hateful posts when hate is referenced or quoted and when swear words are used in non-hateful contexts.
%We primarily compare variance between the predictions of the zero context model and the best-performing Embed \& Concat model, rather than solely against the true labels, in order to assess the influence of incorporating LLM generated context on model behaviour. 

\subsection{Latent Hatred}

It is critical to analyse instances in which the context-enhanced model predicted a false negative while the zero-context model predicted the true positive, demonstrating the potential pitfalls of incorporating additional context. 158 posts (3.6\% of the test set) were misclassified in this way,\footnote{Confusion matrices are provided in Appendix \ref{conf-mat}.} with many such errors occurring when lengthy context is added to explicitly hateful posts. For instance, short posts such as ``\textit{Fight me, communists}'' and ``\textit{Yes kill all Christians}''.\footnote{Generated background context for all example posts is provided in Appendix \ref{examples}, Table \ref{tab:doubtful}.} Adding context may bias the model to treat the combined input as more descriptive than hateful. This is a major issue with context-aware HSD, as extra context does not necessarily make the original post more hateful, meaning classifiers still struggle to recognise it as such.

Conversely, 187 non-hateful posts (4.4\%) were incorrectly classified by the Embed \& Concat system but not by the zero-context one.
Prompting the LLM to relate generated context to potential hate speech may lead to inventing hateful context where none exists, consequently leading to the over-estimation of non-hateful posts as hateful. 
% However, few examples of this could be found in this dataset, especially examples in which the zero-context model predicted correctly and the context-enhanced model did not.
The post ``\textit{this is what becoming a groyper does to your life people, it is an irreconcilable decision}'' is a warning of the damage that falling into right-wing conspiracist groups can do to someone, but is mistakenly classified as hateful once lengthy context about \textit{groypers} \cite{hawleyGroyperMovementUS2021} is added, associating them with hateful ideologies and leading the classifier to over-represent this inferred hate.

\subsection{MAMI}

\begin{figure}[h!]
\centering

\begin{tabular}{@{}p{0.49\linewidth}@{} c p{0.47\linewidth}@{}}
\begin{minipage}[tl]{\linewidth}
    \centering
    \includegraphics[width=0.9\linewidth]{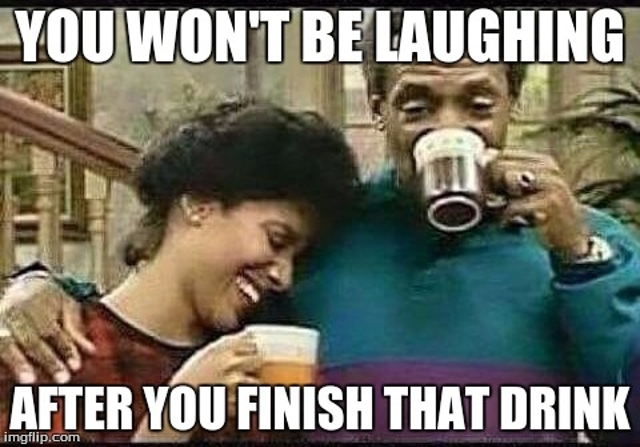}
\end{minipage}
& &
\begin{minipage}[tl]{\linewidth}
    \scriptsize
    ``The meme features a picture of Carl Winslow and his wife Harriette from the sitcom \textit{Family Matters}. The text suggests a scenario where someone is initially happy or carefree ("laughing") but will face negative consequences or regret after consuming the drink. It is likely used to humorously warn against overindulgence or actions that seem fun at the moment but will lead to a negative outcome.''
\end{minipage}
\end{tabular}
\caption{Example false negative meme alongside its LLM-generated context.}
\label{fn}

\end{figure}

68 misogynous memes (3.1\%) were classified correctly with no context but incorrectly by the Embed \& Concat system. These generally fall under the aforementioned issue of adding irrelevant or incorrect context that obfuscates previously clearer hate. 
In the example in Figure \ref{fn}, the generated context does not recognise Bill Cosby in the image and misses the implication of Drug-Facilitated Sexual Assault \cite{butlerDrugfacilitatedSexualAssault2009}, leading the model to misclassify it as not misogynous.
    
\begin{figure}[h!]
\centering

\begin{tabular}{@{}p{0.56\linewidth}@{} c p{0.38\linewidth}@{}}
\begin{minipage}[tl]{\linewidth}
    \centering
    \includegraphics[width=\linewidth]{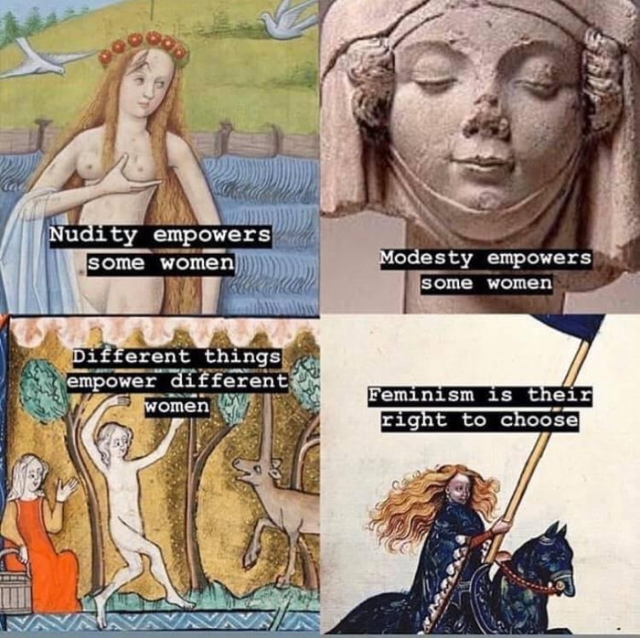}
\end{minipage}
& &
\begin{minipage}[tl]{\linewidth}
    \scriptsize
    ``The meme is a four-panel image featuring historical artwork. Each panel displays a different image of a woman, accompanied by a caption describing an action or state of being that purportedly empowers women. It is likely designed to express the idea that feminism supports a variety of choices for women, as their empowerment may come from diverse actions, behaviours, and lifestyles.''
\end{minipage}
\end{tabular}
\caption{Example false positive meme alongside its LLM-generated context.}
\label{fp}
\end{figure}

Inversely, 63 non-misogynous memes (2.9\%) were mislabelled by the \textit{Embed \& Concat} model but not by the zero-context one. These errors tended to arise from two main phenomena. The first is as before, with context introducing ambiguity or inventing false misogynistic connections. The second is exemplified by the meme in Figure \ref{fp}, which contains many words often used with a negative connotation in misogynistic posts.
Phrases like “modesty”, “feminism”, and “empowerment” are frequently used in polarising or sarcastic contexts, especially in this dataset, potentially leading the model to view supportive posts as insincere.

\section{Discussion}

% Our results suggest that the limitations of previous contextual approaches to HSD stem not only from the availability of background context, but also from how that context is incorporated. 
While prior work in contextualised HSD has relied on entity linking to retrieve information from static knowledge bases, we show that generating background knowledge from the full content of a post consistently led to improved classification accuracy. This supports the hypothesis that contextualisation should be treated as a post-level semantic task rather than an entity-level retrieval task, allowing ambiguous references and implicit meaning to be interpreted within the context of the post.
% The method by which context was incorporated also proved to be crucial. Across datasets and tasks, embedding-level concatenation (Embed \& Concat) consistently outperformed text concatenation, post rewriting, and transformer-based fusion, suggesting that preserving a strict distinction between the original input and generated context helps to prevent semantic interference while still allowing the classifier to exploit auxiliary context.

Comparison with direct LLM classification highlights a trade-off between general reasoning ability and task-specific adaptation. While the LLM classifier achieved strong performance, particularly on binary multimodal classification, our context-enhanced encoder models remained competitive on more nuanced tasks, perhaps as training allows them to better understand the annotation conventions and stylistic expressions of hate speech within a given dataset \cite{albladi-llm-hsd}. Furthermore, separating knowledge generation from classification allows the contribution of context to be analysed independently of the reasoning capabilities of the underlying LLM.

Lastly, while our results demonstrate the promise of using LLM-generated context in HSD, our error analysis highlights some of its pitfalls. We observe that additional context can cause semantic drift in the original post, leading to false positives when benign content was overcontextualised with charged interpretations, and false negatives when hateful intent was neutralised by overly explanatory context.
%Similar issues have been reported in multimodal explanation generation, where generated captions can hallucinate associations, misidentify target groups, or otherwise misinterpret the intended meaning of hateful memes \cite{sood2025memhatecaptioning}.
These findings suggest that generating relevant background knowledge is only part of the challenge; equally important is ensuring that it complements, rather than overrides, the meaning of the original post.

\section{Conclusions}

In this work, we investigated the use of an LLM as a dynamic knowledge base to generate background context and support implicit HSD across textual and textualised multimodal domains. Building on previous work in contextualised HSD, we proposed generating background context from the full content of a post rather than through entity-centric retrieval, and systematically evaluated our generation strategy under four methods of incorporating the resulting knowledge into the input of a downstream classification pipeline.

Our results show that using LLM-generated background context leads to higher HSD performance than both a zero-context baseline and traditional entity-based contextualisation (RQ1), with improvements of up to 3 F1 points in a textual setting and up to 6 F1 points in a multimodal setting over said baseline.
Further, we demonstrate that the method of incorporation makes a meaningful difference to classification (RQ2), finding that embedding-level concatenation (Embed \& Concat) consistently outperformed text concatenation, post rewriting, and transformer-based fusion across tasks. This indicates that preserving a strict distinction between the original input and generated context helps to prevent semantic interference while still allowing the classifier to exploit auxiliary context.
Finally, we show that LLM-generated background knowledge offers a viable alternative to direct LLM classification (RQ3), improving the performance of a lightweight classifier while providing a modular framework for analysing the contribution of contextual information.

More broadly, this work advances contextualised HSD, replacing static entity-centric context retrieval with dynamic, post-specific context generation. As LLMs continue to improve, our results suggest that progress will depend on how context can be generated, validated, and incorporated into the input of downstream classifiers.

\clearpage

\section*{Limitations}

Despite promising results, this research is subject to limitations that should be considered when interpreting the findings or generalising them to broader settings. 
Primarily, we relied on a single LLM (\texttt{Gemini 2.0 Flash}) and a fixed set of simple, natural-language prompts to generate background context. Alternative models, prompt formulations, or temperature settings could produce substantially different context and thus alter downstream performance. Future work could also explore evaluation of the generated context directly, rather than only its effect on downstream performance.

Additionally, our evaluation is limited to two English-language datasets collected over relatively brief time spans. This could limit generalisability to multilingual or non-English settings, as well as temporal generalisability, particularly as the language and references used in online posts and memes changes rapidly and dramatically over time (see \citet{bisera_kostadinovska_stojchevska_2018_1492894} and \citet{joshi2024contextualizinginternetmemessocial} for more on this).
Moreover, human annotations, particularly for implicit hate, are inherently subjective, which may introduce bias into annotated datasets.
Our error analysis revealed several instances that appeared mislabelled or open to multiple valid interpretations in the Latent Hatred dataset. These issues are discussed further in Appendix \ref{doubt}.

%Different annotators' perspectives are also not considered in this case (there are studies that classify HS wrt the annotators' background, etc., that is, hateful for some annotators (depending on their age, cultural background, etc) but not for others) 

Lastly, reliance on macro F1 scores, though useful as an overall performance indicator, can obscure important class-specific trends and failure modes. 
Future evaluations may benefit from more fine-grained analysis or the development of dedicated challenge datasets to better assess robustness to various phenomena, including the aforementioned

\section*{Ethical Considerations}

As argued by \citet{wongWhatSocialBenefit2024}, technical HSD research has advanced rapidly but studies often underemphasise ethical and societal implications, leading to a lack of real-world impact. 
While automatic HSD systems are undoubtedly valuable for identifying and combating hateful content online, their deployment raises substantial ethical concerns that must be addressed alongside technical development.

Firstly and most importantly, HSD systems, if implemented inadequately, can limit freedom of expression, a fundamental right \cite{CoEExpression2025}.
Without frequent assessment and transparent appeal mechanisms, there is a risk of these models being weaponised to suppress lawful content---either inadvertently through biased data and imprecise definitions, or deliberately by actors seeking to limit public discourse on certain topics \cite{RepressivePowerArtificial}.
The ease with which a seemingly robust HSD system can be adapted to identify and suppress any kind of \textit{undesirable} content, merely by retraining it with different labels, should not be underestimated.

In regards to this work more specifically, the use of LLMs to generate background context introduces unique risks. Unlike curated knowledge bases, LLM outputs are probabilistic and not easily verifiable, and may introduce misinformation, stereotypes, or defamatory content directly into the classification pipeline. As our error analysis shows, generated background context can shift the interpretation of posts in unintended ways, leading to false positives or false negatives. In a production setting, such behaviour could unfairly penalise individuals or distort moderation outcomes. These risks underscore the need for mechanisms to validate or fact-check generated context, human review for high-impact decisions, and clear disclaimers about the probabilistic nature of LLM outputs.

% Overall, integrating ethical safeguards from the outset is essential not only to uphold human rights but also to ensure that advances in context-aware HSD achieve genuine social benefit, rather than solely theoretical gains in performance.

% In regards to this work more specifically, one must carefully consider the use of LLMs, especially in their capacity to introduce misleading, incorrect, or biased information in the generated background context.

% As LLM-generated ``knowledge'' is not verifiable in the same way as curated knowledge bases, there is a heightened risk of misinformation entering the moderation pipeline.

\section*{Data and Code Availability}

Processed data, including generated background context, and code to reproduce our experiments are available in an anonymous repository at \url{https://github.com/joshbrook/contextual-hsd}.

\bibliography{references}

@ARTICLE{albladi-llm-hsd,
  author={Albladi, Aish and Islam, Minarul and Das, Amit and Bigonah, Maryam and Zhang, Zheng and Jamshidi, Fatemeh and Rahgouy, Mostafa and Raychawdhary, Nilanjana and Marghitu, Daniela and Seals, Cheryl},
  journal={IEEE Access}, 
  title={Hate Speech Detection Using Large Language Models: A Comprehensive Review}, 
  year={2025},
  volume={13},
  number={},
  pages={20871-20892},
  doi={10.1109/ACCESS.2025.3532397}
}

@inproceedings{bourgeadeHumansNeedContext2024,
    title = {{Humans Need Context, What about Machines? Investigating Conversational Context in Abusive Language Detection}},
    author = "Bourgeade, Tom  and
      Li, Zongmin  and
      Benamara, Farah  and
      Moriceau, V{\'e}ronique  and
      Su, Jian  and
      Sun, Aixin",
    editor = "Calzolari, Nicoletta  and
      Kan, Min-Yen  and
      Hoste, Veronique  and
      Lenci, Alessandro  and
      Sakti, Sakriani  and
      Xue, Nianwen",
    booktitle = "Proceedings of the 2024 Joint International Conference on Computational Linguistics, Language Resources and Evaluation (LREC-COLING 2024)",
    month = may,
    year = "2024",
    address = "Torino, Italia",
    publisher = "ELRA and ICCL",
    url = "https://aclanthology.org/2024.lrec-main.740/",
    pages = "8438--8452",
}

@article{butlerDrugfacilitatedSexualAssault2009,
  title = {Drug-Facilitated Sexual Assault},
  author = {Butler, Bernadette and Welch, Jan},
  year = {2009},
  month = mar,
  journal = {CMAJ : Canadian Medical Association Journal},
  volume = {180},
  number = {5},
  pages = {493--494},
  issn = {0820-3946},
  doi = {10.1503/cmaj.090006},
  urldate = {2025-06-18},
  pmcid = {PMC2645469},
  pmid = {19255067}
}

@inproceedings{lmarena,
author = {Chiang, Wei-Lin and Zheng, Lianmin and Sheng, Ying and Angelopoulos, Anastasios N. and Li, Tianle and Li, Dacheng and Zhu, Banghua and Zhang, Hao and Jordan, Michael I. and Gonzalez, Joseph E. and Stoica, Ion},
title = {Chatbot arena: an open platform for evaluating {LLMs} by human preference},
year = {2024},
publisher = {JMLR.org},
booktitle = {Proceedings of the 41st International Conference on Machine Learning},
articleno = {331},
numpages = {30},
location = {Vienna, Austria},
series = {ICML'24},
url = {https://dl.acm.org/doi/abs/10.5555/3692070.3692401}
}

@misc{chiu2022detectinghatespeechgpt3,
      title={{Detecting Hate Speech with GPT-3}}, 
      author={Ke-Li Chiu and Annie Collins and Rohan Alexander},
      year={2022},
      eprint={2103.12407},
      archivePrefix={arXiv},
      primaryClass={cs.CL},
      url={https://arxiv.org/abs/2103.12407}, 
}

@inproceedings{elsheriefLatentHatred2021,
  title = {Latent {{Hatred}}: {{A Benchmark}} for {{Understanding Implicit Hate Speech}}},
  shorttitle = {Latent {{Hatred}}},
  booktitle = {Proceedings of the 2021 {{Conference}} on {{Empirical Methods}} in {{Natural Language Processing}}},
  author = {ElSherief, Mai and Ziems, Caleb and Muchlinski, David and Anupindi, Vaishnavi and Seybolt, Jordyn and De Choudhury, Munmun and Yang, Diyi},
  editor = {Moens, Marie-Francine and Huang, Xuanjing and Specia, Lucia and Yih, Scott Wen-tau},
  year = {2021},
  month = nov,
  pages = {345--363},
  publisher = {Association for Computational Linguistics},
  address = {Online and Punta Cana, Dominican Republic},
  doi = {10.18653/v1/2021.emnlp-main.29},
  urldate = {2025-02-26}
}

@inproceedings{fersiniMAMI2022,
  title = {{{SemEval-2022 Task}} 5: {{Multimedia Automatic Misogyny Identification}}},
  shorttitle = {{{SemEval-2022 Task}} 5},
  booktitle = {Proceedings of the 16th {{International Workshop}} on {{Semantic Evaluation}} ({{SemEval-2022}})},
  author = {Fersini, Elisabetta and Gasparini, Francesca and Rizzi, Giulia and Saibene, Aurora and Chulvi, Berta and Rosso, Paolo and Lees, Alyssa and Sorensen, Jeffrey},
  editor = {Emerson, Guy and Schluter, Natalie and Stanovsky, Gabriel and Kumar, Ritesh and Palmer, Alexis and Schneider, Nathan and Singh, Siddharth and Ratan, Shyam},
  year = {2022},
  month = jul,
  pages = {533--549},
  publisher = {Association for Computational Linguistics},
  address = {Seattle, United States},
  doi = {10.18653/v1/2022.semeval-1.74},
  urldate = {2025-05-16}
}

@inproceedings{gaoDetectingOnlineHate2017,
  title = {Detecting {{Online Hate Speech Using Context Aware Models}}},
  booktitle = {Proceedings of the {{International Conference Recent Advances}} in {{Natural Language Processing}}, {{RANLP}} 2017},
  author = {Gao, Lei and Huang, Ruihong},
  year = {2017},
  pages = {260--266},
  publisher = {INCOMA Ltd.},
  doi = {10.26615/978-954-452-049-6_036}
}

@misc{greif2025multimodalllmsocrocr,
      title={{Multimodal LLMs for OCR, OCR Post-Correction, and Named Entity Recognition in Historical Documents}}, 
      author={Gavin Greif and Niclas Griesshaber and Robin Greif},
      year={2025},
      eprint={2504.00414},
      archivePrefix={arXiv},
      primaryClass={cs.CL},
      url={https://arxiv.org/abs/2504.00414}, 
}

@misc{guoInvestigationLargeLanguage2024,
  title = {An {{Investigation}} of {{Large Language Models}} for {{Real-World Hate Speech Detection}}},
  author = {Guo, Keyan and Hu, Alexander and Mu, Jaden and Shi, Ziheng and Zhao, Ziming and Vishwamitra, Nishant and Hu, Hongxin},
  year = {2024},
  month = jan,
  number = {arXiv:2401.03346},
  eprint = {2401.03346},
  primaryclass = {cs},
  publisher = {arXiv},
  doi = {10.48550/arXiv.2401.03346},
  urldate = {2025-02-27},
  archiveprefix = {arXiv}
}

@incollection{hawleyGroyperMovementUS2021,
  title = {The ``{{Groyper}}'' Movement in the {{US}}: {{Challenges}} for the Post-{{Alt-right}}},
  shorttitle = {The ``{{Groyper}}'' Movement in the {{US}}},
  booktitle = {Contemporary {{Far-Right Thinkers}} and the {{Future}} of {{Liberal Democracy}}},
  author = {Hawley, George},
  year = {2021},
  publisher = {Routledge},
  isbn = {978-1-003-10517-6},
  url = {https://www.taylorfrancis.com/chapters/edit/10.4324/9781003105176-17/groyper-movement-us-george-hawley}
}

@inproceedings{hendersonHowDogwhistlesWork2018,
  title = {How {{Dogwhistles Work}}},
  booktitle = {New {{Frontiers}} in {{Artificial Intelligence}}},
  author = {Henderson, R. and McCready, Elin},
  editor = {Arai, Sachiyo and Kojima, Kazuhiro and Mineshima, Koji and Bekki, Daisuke and Satoh, Ken and Ohta, Yuiko},
  year = {2018},
  pages = {231--240},
  publisher = {Springer International Publishing},
  address = {Cham},
  doi = {10.1007/978-3-319-93794-6_16},
  isbn = {978-3-319-93794-6},
  langid = {english},
}

@misc{kiela2021hatefulmemeschallengedetecting,
      title={The Hateful Memes Challenge: Detecting Hate Speech in Multimodal Memes}, 
      author={Douwe Kiela and Hamed Firooz and Aravind Mohan and Vedanuj Goswami and Amanpreet Singh and Pratik Ringshia and Davide Testuggine},
      year={2021},
      eprint={2005.04790},
      archivePrefix={arXiv},
      primaryClass={cs.AI},
      url={https://arxiv.org/abs/2005.04790}, 
}

@inproceedings{MAMIHakimovCE22,
  author    = {Sherzod Hakimov and
               Gullal Singh Cheema and
               Ralph Ewerth},
  editor    = {Guy Emerson and
               Natalie Schluter and
               Gabriel Stanovsky and
               Ritesh Kumar and
               Alexis Palmer and
               Nathan Schneider and
               Siddharth Singh and
               Shyam Ratan},
  title     = {{TIB-VA} at SemEval-2022 Task 5: {A} Multimodal Architecture for the
               Detection and Classification of Misogynous Memes},
  booktitle = {Proceedings of the 16th International Workshop on Semantic Evaluation,
               SemEval@NAACL 2022, Seattle, Washington, United States, July 14-15,
               2022},
  pages     = {756--760},
  publisher = {Association for Computational Linguistics},
  year      = {2022},
  doi       = {10.18653/v1/2022.semeval-1.105},
}

@misc{linLeveragingWorldKnowledge2022,
  title = {Leveraging {{World Knowledge}} in {{Implicit Hate Speech Detection}}},
  author = {Lin, Jessica},
  year = {2022},
  month = dec,
  number = {arXiv:2212.14100},
  eprint = {2212.14100},
  primaryclass = {cs},
  publisher = {arXiv},
  doi = {10.48550/arXiv.2212.14100},
  urldate = {2025-02-26},
  archiveprefix = {arXiv},
}

@inproceedings{pavlopoulos-2020-toxicity,
    title = "Toxicity Detection: Does Context Really Matter?",
    author = "Pavlopoulos, John  and
      Sorensen, Jeffrey  and
      Dixon, Lucas  and
      Thain, Nithum  and
      Androutsopoulos, Ion",
    editor = "Jurafsky, Dan  and
      Chai, Joyce  and
      Schluter, Natalie  and
      Tetreault, Joel",
    booktitle = "Proceedings of the 58th Annual Meeting of the Association for Computational Linguistics",
    month = jul,
    year = "2020",
    address = "Online",
    publisher = "Association for Computational Linguistics",
    doi = "10.18653/v1/2020.acl-main.396",
    pages = "4296--4305"
}

@inproceedings{markovRoleContextDetecting2022,
  title = {The {{Role}} of {{Context}} in {{Detecting}} the {{Target}} of {{Hate Speech}}},
  booktitle = {Proceedings of the {{Third Workshop}} on {{Threat}}, {{Aggression}} and {{Cyberbullying}} ({{TRAC}} 2022)},
  author = {Markov, Ilia and Daelemans, Walter},
  year = {2022},
  pages = {37--42},
  publisher = {Association for Computational Linguistics},
  url = {https://aclanthology.org/2022.trac-1.5/}
}

@misc{meniniAbuseContextualWhat2021,
  title = {Abuse Is {{Contextual}}, {{What}} about {{NLP}}? {{The Role}} of {{Context}} in {{Abusive Language Annotation}} and {{Detection}}},
  shorttitle = {Abuse Is {{Contextual}}, {{What}} about {{NLP}}?},
  author = {Menini, Stefano and Aprosio, Alessio Palmero and Tonelli, Sara},
  year = {2021},
  month = mar,
  number = {arXiv:2103.14916},
  eprint = {2103.14916},
  primaryclass = {cs},
  publisher = {arXiv},
  doi = {10.48550/arXiv.2103.14916},
  urldate = {2025-02-27},
  archiveprefix = {arXiv},
}

@inproceedings{ocampo-etal-2023-depth,
    title = "An In-depth Analysis of Implicit and Subtle Hate Speech Messages",
    author = "Ocampo, Nicol{\'a}s Benjam{\'i}n  and
      Sviridova, Ekaterina  and
      Cabrio, Elena  and
      Villata, Serena",
    editor = "Vlachos, Andreas  and
      Augenstein, Isabelle",
    booktitle = "Proceedings of the 17th Conference of the European Chapter of the Association for Computational Linguistics",
    month = may,
    year = "2023",
    address = "Dubrovnik, Croatia",
    publisher = "Association for Computational Linguistics",
    url = "https://aclanthology.org/2023.eacl-main.147/",
    doi = "10.18653/v1/2023.eacl-main.147",
    pages = "1997--2013"
}

@misc{pandianiToxicMemesSurvey2024,
  title = {Toxic {{Memes}}: {{A Survey}} of {{Computational Perspectives}} on the {{Detection}} and {{Explanation}} of {{Meme Toxicities}}},
  shorttitle = {Toxic {{Memes}}},
  author = {Pandiani, Delfina Sol Martinez and Sang, Erik Tjong Kim and Ceolin, Davide},
  year = {2024},
  month = jun,
  number = {arXiv:2406.07353},
  eprint = {2406.07353},
  primaryclass = {cs},
  publisher = {arXiv},
  doi = {10.48550/arXiv.2406.07353},
  urldate = {2025-02-27},
  archiveprefix = {arXiv}
}

@article{perezAssessingImpactContextual2023,
  title = {Assessing the {{Impact}} of {{Contextual Information}} in {{Hate Speech Detection}}},
  author = {P{\'e}rez, Juan Manuel and Luque, Franco M. and Zayat, Demian and Kondratzky, Mart{\'i}n and Moro, Agust{\'i}n and Serrati, Pablo Santiago and Zajac, Joaqu{\'i}n and Miguel, Paula and Debandi, Natalia and Gravano, Agust{\'i}n and Cotik, Viviana},
  year = {2023},
  journal = {IEEE Access},
  volume = {11},
  pages = {30575--30590},
  doi = {10.1109/ACCESS.2023.3258973}
}

@misc{petroniLLMsAsKBs2019,
  title = {Language {{Models}} as {{Knowledge Bases}}?},
  author = {Petroni, Fabio and Rockt{\"a}schel, Tim and Lewis, Patrick and Bakhtin, Anton and Wu, Yuxiang and Miller, Alexander H. and Riedel, Sebastian},
  year = {2019},
  month = sep,
  number = {arXiv:1909.01066},
  eprint = {1909.01066},
  primaryclass = {cs},
  publisher = {arXiv},
  doi = {10.48550/arXiv.1909.01066},
  urldate = {2025-02-28},
  archiveprefix = {arXiv}
}

@inproceedings{royProbingLLMsHate2023,
  title = {Probing {{LLMs}} for Hate Speech Detection: Strengths and Vulnerabilities},
  shorttitle = {Probing {{LLMs}} for Hate Speech Detection},
  booktitle = {Findings of the {{Association}} for {{Computational Linguistics}}: {{EMNLP}} 2023},
  author = {Roy, Sarthak and Harshvardhan, Ashish and Mukherjee, Animesh and Saha, Punyajoy},
  editor = {Bouamor, Houda and Pino, Juan and Bali, Kalika},
  year = {2023},
  month = dec,
  pages = {6116--6128},
  publisher = {Association for Computational Linguistics},
  address = {Singapore},
  doi = {10.18653/v1/2023.findings-emnlp.407},
  urldate = {2025-02-27}
}

@inproceedings{sharifiradBoostingTextClassification2018,
  title = {Boosting {{Text Classification Performance}} on {{Sexist Tweets}} by {{Text Augmentation}} and {{Text Generation Using}} a {{Combination}} of {{Knowledge Graphs}}},
  booktitle = {Proceedings of the 2nd {{Workshop}} on {{Abusive Language Online}} ({{ALW2}})},
  author = {Sharifirad, Sima and Jafarpour, Borna and Matwin, Stan},
  editor = {Fi{\v s}er, Darja and Huang, Ruihong and Prabhakaran, Vinodkumar and Voigt, Rob and Waseem, Zeerak and Wernimont, Jacqueline},
  year = {2018},
  month = oct,
  pages = {107--114},
  publisher = {Association for Computational Linguistics},
  address = {Brussels, Belgium},
  doi = {10.18653/v1/W18-5114},
  urldate = {2025-02-27}
}

@inproceedings{sood-memhatecaptioning,
author = {Sood, Rishik and Anaissi, Ali and Huang, Weidong and Braytee, Ali},
title = {{MemHateCaptioning: Enhancing Hate Speech Detection in Memes with Context-Aware Captioning and Chain-of-Thought}},
year = {2025},
isbn = {9798400713316},
publisher = {Association for Computing Machinery},
address = {New York, NY, USA},
url = {https://doi.org/10.1145/3701716.3718385},
doi = {10.1145/3701716.3718385},
booktitle = {Companion Proceedings of the ACM on Web Conference 2025},
pages = {2034–2041},
numpages = {8},
location = {Sydney NSW, Australia},
series = {WWW '25}
}

@misc{wangKGsFromLLMs2020,
  title = {Language {{Models}} Are {{Open Knowledge Graphs}}},
      author={Chenguang Wang and Xiao Liu and Dawn Song},
      year={2020},
      eprint={2010.11967},
      archivePrefix={arXiv},
      primaryClass={cs.CL},
      url={https://arxiv.org/abs/2010.11967}, 
}

@inproceedings{vanHulstREL2020,
 author =    {van Hulst, Johannes M. and Hasibi, Faegheh and Dercksen, Koen and Balog, Krisztian and de Vries, Arjen P.},
 title =     {{REL: An Entity Linker Standing on the Shoulders of Giants}},
 booktitle = {Proceedings of the 43rd International ACM SIGIR Conference on Research and Development in Information Retrieval},
 series =    {SIGIR '20},
 year =      {2020},
 publisher = {ACM},
   url={http://dx.doi.org/10.1145/3397271.3401416},
}

@article{speer2017conceptnet,
  author       = {Robyn Speer and
                  Joshua Chin and
                  Catherine Havasi},
  title        = {ConceptNet 5.5: An Open Multilingual Graph of General Knowledge},
  journal      = {CoRR},
  year         = {2016},
  url          = {http://arxiv.org/abs/1612.03975},
  eprinttype    = {arXiv},
  eprint       = {1612.03975},
  bibsource    = {dblp computer science bibliography, https://dblp.org}
}

@inproceedings{reimers-2019-sentence-bert,
  title = {{Sentence-BERT: Sentence Embeddings using Siamese BERT-Networks}},
  author = "Reimers, Nils and Gurevych, Iryna",
  booktitle = "Proceedings of the 2019 Conference on Empirical Methods in Natural Language Processing",
  month = "11",
  year = "2019",
  publisher = "Association for Computational Linguistics",
  url = "https://arxiv.org/abs/1908.10084",
}

@misc{radford2021CLIP,
      title={Learning Transferable Visual Models From Natural Language Supervision}, 
      author={Alec Radford and Jong Wook Kim and Chris Hallacy and Aditya Ramesh and Gabriel Goh and Sandhini Agarwal and Girish Sastry and Amanda Askell and Pamela Mishkin and Jack Clark and Gretchen Krueger and Ilya Sutskever},
      year={2021},
      eprint={2103.00020},
      archivePrefix={arXiv},
      primaryClass={cs.CV},
      url={https://arxiv.org/abs/2103.00020}, 
}

@article{roberta,
  author       = {Yinhan Liu and
                  Myle Ott and
                  Naman Goyal and
                  Jingfei Du and
                  Mandar Joshi and
                  Danqi Chen and
                  Omer Levy and
                  Mike Lewis and
                  Luke Zettlemoyer and
                  Veselin Stoyanov},
  title        = {{RoBERTa}: {A} Robustly Optimized {BERT} Pretraining Approach},
  journal      = {CoRR},
  volume       = {abs/1907.11692},
  year         = {2019},
  url          = {http://arxiv.org/abs/1907.11692},
  eprinttype    = {arXiv},
  eprint       = {1907.11692},
  bibsource    = {dblp computer science bibliography, https://dblp.org}
}

@misc{fountaLargeScaleCrowdsourcing2018,
  title = {Large {{Scale Crowdsourcing}} and {{Characterization}} of {{Twitter Abusive Behavior}}},
  author = {Founta, Antigoni-Maria and Djouvas, Constantinos and Chatzakou, Despoina and Leontiadis, Ilias and Blackburn, Jeremy and Stringhini, Gianluca and Vakali, Athena and Sirivianos, Michael and Kourtellis, Nicolas},
  year = {2018},
  month = apr,
  number = {arXiv:1802.00393},
  eprint = {1802.00393},
  primaryclass = {cs},
  publisher = {arXiv},
  doi = {10.48550/arXiv.1802.00393},
  urldate = {2025-02-28},
  archiveprefix = {arXiv}
}

@misc{bert-devlin-2019,
      title={{BERT: Pre-training of Deep Bidirectional Transformers for Language Understanding}}, 
      author={Jacob Devlin and Ming-Wei Chang and Kenton Lee and Kristina Toutanova},
      year={2019},
      eprint={1810.04805},
      archivePrefix={arXiv},
      primaryClass={cs.CL},
      url={https://arxiv.org/abs/1810.04805}, 
}

@inproceedings{doganc-markov-2023-generic,
    title = "From Generic to Personalized: Investigating Strategies for Generating Targeted Counter Narratives against Hate Speech",
    author = "Do{\u{g}}an{\c{c}}, Mekselina  and
      Markov, Ilia",
    editor = "Chung, Yi-Ling  and
      Bonaldi, Helena  and
      Abercrombie, Gavin  and
      Guerini, Marco",
    booktitle = "Proceedings of the 1st Workshop on CounterSpeech for Online Abuse (CS4OA)",
    month = sep,
    year = "2023",
    address = "Prague, Czechia",
    publisher = "Association for Computational Linguistics",
    url = "https://aclanthology.org/2023.cs4oa-1.1/",
    pages = "1--12",
}

@misc{wongWhatSocialBenefit2024,
  title = {What Is the Social Benefit of Hate Speech Detection Research? {{A Systematic Review}}},
  shorttitle = {What Is the Social Benefit of Hate Speech Detection Research?},
  author = {Wong, Sidney Gig-Jan},
  year = {2024},
  month = sep,
  number = {arXiv:2409.17467},
  eprint = {2409.17467},
  primaryclass = {cs},
  publisher = {arXiv},
  doi = {10.48550/arXiv.2409.17467},
  urldate = {2025-02-27},
  archiveprefix = {arXiv}
}

@misc{CoEExpression2025,
  author       = "{Council of Europe}",
  title        = "{Freedom of Expression}",
  year         = "2025",
  note         = "Accessed 6 June 2025",
  url = {https://www.coe.int/en/web/echr-toolkit/la-liberte-dexpression}
}

@article{RepressivePowerArtificial,
  title = {The {{Repressive Power}} of {{Artificial Intelligence}}},
  journal = {Freedom House},
  year = {2023},
  author = {Allie Funk and Adrian Shahbaz and Kian Vesteinsson},
  url = {https://freedomhouse.org/report/freedom-net/2023/repressive-power-artificial-intelligence},
}

@article{bisera_kostadinovska_stojchevska_2018_1492894,
  author       = {Bisera Kostadinovska-Stojchevska and
                  Elena Shalevska},
  title        = {INTERNET MEMES AND  THEIR SOCIO-LINGUISTIC
                   FEATURES
                  },
  journal      = {European Journal of Literature, Language and
                   Linguistics Studies
                  },
  year         = 2018,
  volume       = 2,
  number       = 4,
  month        = nov,
  doi          = {10.5281/zenodo.1492894},
  url          = {https://doi.org/10.5281/zenodo.1492894},
}

@misc{joshi2024contextualizinginternetmemessocial,
      title={Contextualizing Internet Memes Across Social Media Platforms}, 
      author={Saurav Joshi and Filip Ilievski and Luca Luceri},
      year={2024},
      eprint={2311.11157},
      archivePrefix={arXiv},
      primaryClass={cs.SI},
      url={https://arxiv.org/abs/2311.11157}, 
}

@inproceedings{britez2025cltl,
  title     = {{CLTL at EXIST 2025: Identifying Sexist Memes Using an Ensemble of Shallow and Transformer Models}},
  author    = {Britez, Ariana and Markov, Ilia},
  booktitle = {CLEF 2025 Working Notes},
  year      = {2025},
  address   = {Madrid, Spain},
  month     = sep,
  publisher = {CEUR Workshop Proceedings},
  url       = {https://ceur-ws.org/Vol-4038/paper_143.pdf},
  note      = {Notebook for the EXIST Lab at CLEF 2025},
}

@misc{das2020detectinghatespeechmultimodal,
      title={Detecting Hate Speech in Multi-modal Memes}, 
      author={Abhishek Das and Japsimar Singh Wahi and Siyao Li},
      year={2020},
      eprint={2012.14891},
      archivePrefix={arXiv},
      primaryClass={cs.CV},
      url={https://arxiv.org/abs/2012.14891}, 
}

@misc{wikidata,
  author = {{Wikidata}},
  title = {Wikidata: A free knowledge base that anyone can edit},
  year = {2025},
  url = {https://www.wikidata.org},
  note = {Accessed: 2025-04-11}
}

@misc{gemini2flash,
  author = {Google DeepMind},
  title = {Gemini 2.0 Flash},
  year = {2024},
  url = {https://ai.google.dev/gemini-api/docs/models/gemini-2.0-flash},
  note = {Accessed: 2025-06-12}
}

\appendix

\clearpage

\section*{Appendix}
\renewcommand{\thesubsection}{\Alph{subsection}}

\subsection{A Note on the Radboud Entity Linker} \label{rel-note}

Our replication of \citeposs{linLeveragingWorldKnowledge2022} methodology was unable to validate their results, instead demonstrating a slight drop in macro F1 score from the zero-context baseline for both binary and multi-class classification.
One major issue with Lin's approach is REL itself, which struggles to identify entities that are not capitalised in the ``proper'' way --- a substantial issue when working with informal social media data. The paper does not explicitly mention the number of entities discovered, but our reproduction finds that only 9,595 of 21,480 posts (44.67\%) were linked to one or more entities.
Manual inspection of a random sample of 100 posts (containing 79 identified tags) found 28 extra named entities missed by REL in Lin's approach, along with 18 incorrectly linked named entities.

\subsection{NER in a Multimodal Setting} \label{no-nes}

Named Entity Recognition (NER) (and by extension, entity linking) on multimodal input, as in the MAMI dataset, is ill-defined. Even if named entities are present, they may appear either directly in the embedded text or visually within the image.
Running the REL model on the extracted texts identified 6,598 entities spanning 3,394 of the 10,995 memes, meaning only 31\% of memes have at least one entity with an average of 0.6 entities per meme, significantly lower than in the textual Latent Hatred set.
``\textit{trump}'', ``\textit{china}'', and ``\textit{covid}'' appear frequently in these meme texts, demonstrating the time and political context from which they were originally collated.
Running REL on the generated image descriptions reveals another 4,478 named entities. However, given that memes generally rely on specific formats, often based on stills from movies or TV shows, the majority of these entities are likely irrelevant to detecting hate speech. Demonstrating this, the most frequently occurring entities found were ``\textit{spongebob squarepants}'', ``\textit{thanos}'', and ``\textit{gordon ramsay}'', showing both the kinds of references used and the high identifiability of these characters to our image captioning model.\footnote{Some examples of memes with potentially irrelevant named entities are given in Appendix \ref{examples}, Figure \ref{fig:irrel-nes}.}
In general, we find a lack of reliable entity anchors and suggest that, while it makes mores sense to identify entities from the extracted text, recall is still low.
% Additionally, our proposed prompting strategy generally subsumes NER. As LLMs are used to generate background context from the image, these texts should inherently cover any entity-relevant content that might otherwise be retrieved through NER.

\subsection{Representation of Multimodal Input} \label{representation}

In this study, we represent memes entirely through text. We stress, primarily, that this leads to improved interpretability, as text-based inputs can be directly inspected for error or bias. 
In contrast, visual embeddings (from e.g., CLIP \cite{radford2021CLIP}), which have been used previously in research involving this dataset (by e.g., \citet{MAMIHakimovCE22}) can be opaque, offering limited insight into the basis for a model's decisions. Secondarily, this approach reduces in-house computational cost and allows for the reuse of familiar tools without specialised multimodal infrastructure.
The inclusion of the image description is, of course, crucial. The meaning of a meme can generally only be derived from the combination of textual and visual modalities, and captions provide semantic grounding that would be lost by using only the extracted text.

Manual investigation of the extracted meme texts provided with the original MAMI data also revealed some inconsistencies (e.g., transcription errors, unclear spacing, and including watermarks and timestamps), prompting us to re-process the whole dataset.
\citet{greif2025multimodalllmsocrocr} demonstrate that multimodal LLMs can surpass the accuracy of conventional optical character recognition tools, leading us to again use the \texttt{Gemini 2.0 Flash} model to generate these new representations.

\subsection{Doubtful Labels in Latent Hatred} \label{doubt}

Through performing an error analysis on the Latent Hatred dataset, we discovered a substantial number of posts with doubtful labels. 
For instance, ``\textit{Berlin is the new Jerusalem: more and more Israeli Jews moving to Germany \#swrm via ...}'' is labelled as \textit{hate} and \textit{stereotypical}, but the reasoning behind this is unclear. Presumably, this is just a tweet linking to a news article, and does not appear to be hateful, instead just reporting a trend.
Potentially inaccurate or misleading annotations are particularly noticeable in the dataset's multi-class labels, where each instance is constrained to a single category despite often exhibiting characteristics of multiple classes. 
For instance, many tweets containing general references to race and whiteness (e.g., ``\textit{\#aryan men \& women will either rise up to fight for their heritage, children \& future or the white race will cease to exist. choose now.}'') are labelled as \textit{white grievance}, while they may better suit another class (\textit{incitement} in this case).
% This prevalence of doubtful labels in hate speech datasets has been observed in various previous studies \cite{van-aken-etal-2018-challenges, markov-ensemble}, and is often proposed as a leading cause of lower-than-expected model performance.

\begin{comment}
    
\subsection{Differences in Baseline Results on MAMI and Latent Hatred} \label{diffs}

Higher baseline performance on the MAMI dataset compared to Latent Hatred can largely be attributed to differences in the nature of the tasks and data. Latent Hatred involves implicit HSD, which is more inherently subtle, ambiguous, and context-dependent. In contrast, MAMI focuses on misogynous memes, which, while multimodal, meaning their hate can generally only be understood based on a union of modalities, are generally more explicit in their intent. The hateful post is sometimes also visually reinforced or paired with more overtly stereotypical language, making it easier for classifiers to detect, especially when background context can bridge the modality gap. Moreover, memes tend to stand alone, structured to communicate a single idea in isolation, whereas posts from Latent Hatred have been occasionally extracted mid-conversation, losing surrounding conversational context that may help to clarify intent.
Additionally, Latent Hatred was specifically curated to contain examples that are ambiguous and difficult to predict, while MAMI acts as a more generalist dataset. As such, even simple models can perform better on MAMI at baseline, while struggling with the more oblique content found in Latent Hatred.

\end{comment}

\clearpage

\onecolumn

\subsection{LLM Prompts for Context Generation, Incorporation, and HSD} \label{tab:prompts}

\begin{table}[ht!]
\footnotesize
\centering
\setlength{\tabcolsep}{5pt}
\begin{tabularx}{\textwidth}{@{}p{5cm}p{8cm}X@{}}

\toprule
\textbf{System} & \textbf{Request} & \textbf{Input(s)} \\
\midrule

You are a system that extracts text from memes.
& Extract and return only the text directly from this image. Ignore any watermarks. Respond with just the extracted text.
& Meme (JPEG) \\
\midrule

You are a system that generates descriptions of images.
& In one sentence, describe who or what is in this image and what they are doing. Ignore all text.
& Meme (JPEG) \\
\midrule

You are a system that finds and adds background context to tweets. You respond briefly with a plaintext paragraph and stick to the facts.
& Give background context about the following tweet. You should identify any implicit meaning or sarcasm, provide some information on any obscure named entities, and give a general overview of any relevant world knowledge to make the tweet easier to understand.
& Post \\
\midrule

You are a system that finds and adds background context to memes. You respond briefly with a plaintext paragraph and stick to the facts.
& Give background context about the following meme. You should identify any implicit meaning or sarcasm, provide some information on any obscure named entities, and give a general overview of any relevant world knowledge to make the meme easier to understand.
& Meme (JPEG) \\
\midrule

You are a system that incorporates background context context into tweets.
& Given the following tweet and its context, incorporate this context back into the original tweet. Keep the tweet's original structure and intent intact while embedding additional information. Respond with only the newly-generated tweet. 
& \makecell[tl]{Post,\\ Context} \\
\midrule

You are a system that incorporates background context into textual representations of memes. 
& The following text has been extracted directly from a meme using OCR, and also includes a brief description of the meme's image and some background context. Incorporate these into one unified text that represents the meme and its context. Keep some of the meme text's original structure and intent intact while embedding additional information. Respond with only the newly-generated textual meme representation. 
& \makecell[tl]{Extracted\\Text,\\Image\\Description,\\Context}\\

\midrule

You are a system that detects hate speech in tweets.
& Is the following tweet considered hate speech? Respond with one word: 'yes' or 'no'.
& Post \\

\midrule

You are a system that classifies implicitly hateful tweets into the following classes. You respond only with the name of one class.
& Which of the following classes does this tweet best fit into? ('White Grievance', 'Incitement', 'Stereotypical', 'Inferiority', 'Irony', Threatening', 'Other').
& Post \\

\midrule

You are a system that detects misogyny in memes.
& Is this meme misogynous? Respond only with one word: 'yes' or 'no'.
& Meme (JPEG) \\

\midrule

You are a system performs multi-label classification of misogynous memes. You respond only with the names of all suitable classes.
& Which of the following classes does this misogynous meme fit into? ('Shaming', 'Stereotype', 'Objectification', 'Violence'). If it is not misogynous, respond 'None'.
& Meme (JPEG) \\

\bottomrule
\end{tabularx}
\caption{System, request, and input(s) for HSD-related prompts.}

\end{table}

\clearpage

\subsection{Dataset Statistics} \label{dists}

\begin{table}[h]
\footnotesize
\centering
\begin{tabular}{l>{\centering\arraybackslash}c>{\centering\arraybackslash}c>
                {\centering\arraybackslash}c>{\centering\arraybackslash}c>
                {\centering\arraybackslash}c>{\centering\arraybackslash}c>
                {\centering\arraybackslash}c}
    \toprule
    & \makecell[cc]{White\\ Grievance} & Incitement & \makecell[cc]{Stereo-\\ typing} & Inferiority & Irony & \makecell[cc]{Threat-\\ ening} & Other \\
    \midrule
    Count & 1,538 & 1,269 & 1,133 & 863 & 797 & 666 & 80 \\
    Occurrence & 24.24\% & 19.99\% & 17.85\% & 13.60\% & 12.56\% & 10.49\% & 1.26\% \\
    \bottomrule
\end{tabular}
\caption{Distribution of the Latent Hatred dataset's implicit hate classes.}
\label{tab:class_dist}
\end{table}

\begin{table}[h!]
\footnotesize
    \centering
    \begin{tabular}{lcccc}
        \toprule
        & Stereotyping & Objectification & Shaming & Violence \\
        \midrule
        Count & 3,160 & 2,549 & 1,417 & 1,106 \\
        Occurrence & 57.46\% & 46.35\% & 25.76\% & 20.11\% \\
        \bottomrule
    \end{tabular}
    \caption{Distribution of misogynous labels in the MAMI dataset.}
    \label{label_dist}
\end{table}

\subsection{Example posts, memes, and generated context} \label{examples}

\setlength{\parindent}{0pt}
\setlength{\intextsep}{12pt}
\setlength{\textfloatsep}{12pt}

\begin{figure}[H]
\label{fig:ex-memes}
  \centering
  
  \begin{subfigure}[b]{0.31\textwidth}
    \centering
    \subcaption{Not Misogynous\\ (No Sub-Label)}
    \includegraphics[width=\linewidth, valign=m]{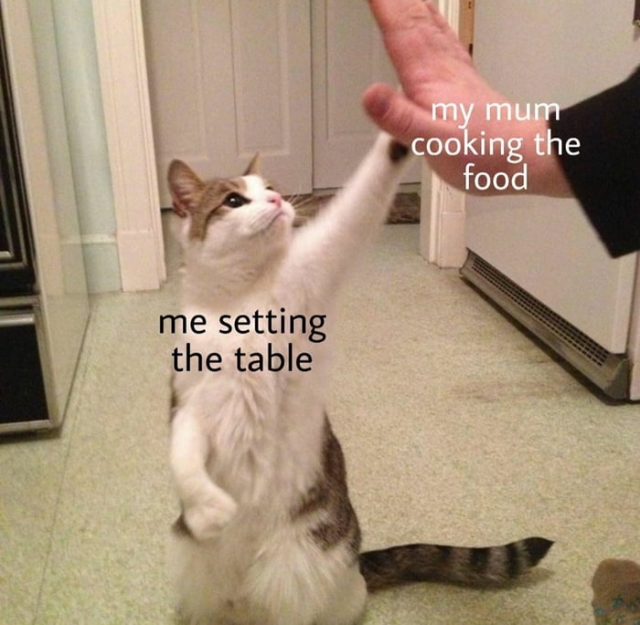}
    
  \end{subfigure}
  \hfill
  \begin{subfigure}[b]{0.32\textwidth}
    \centering
    \subcaption{Misogynous\\ (violence)}
    \includegraphics[width=\linewidth, valign=m]{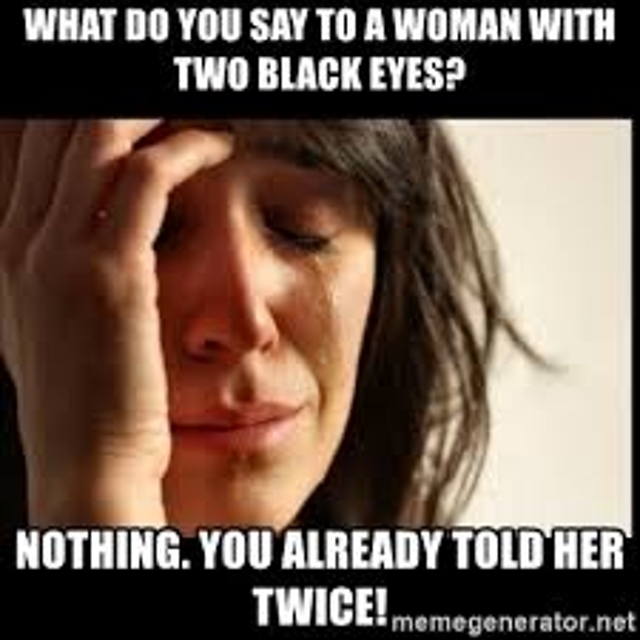}
    
  \end{subfigure}
  \hfill
  \begin{subfigure}[b]{0.33\textwidth}
    \centering
    \subcaption{Misogynous\\ (stereotype, objectification)}
    \includegraphics[width=0.8\linewidth, valign=m]{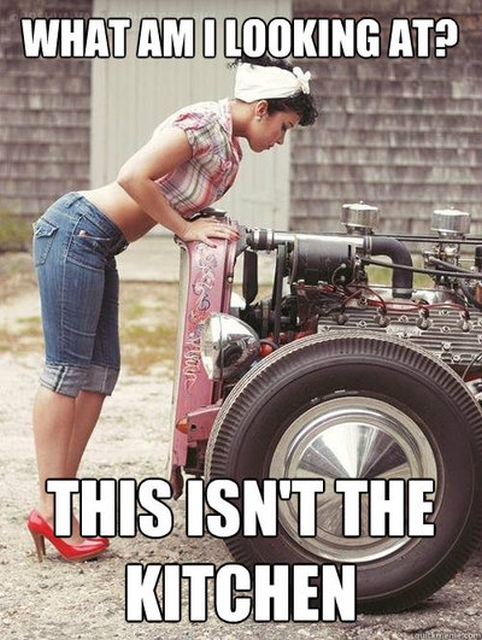}
    
  \end{subfigure}
  
  \caption{Examples of memes and their labels from the MAMI dataset.}
  \label{example-memes}
\end{figure}

\begin{table}[h]
    \centering
    \footnotesize
    \begin{tabularx}{\textwidth}{p{2.5cm}X}
        \toprule
        \textbf{Post} & ``report \textbf{trump} sought visas to import at least 1100 foreign workers into \textbf{usa}'' \\
        \midrule
        \makecell[tl]{\textbf{Entity-Linked}\\ \textbf{Context (REL)}} 
        & ``A trump is a playing card which is elevated above its usual rank in trick-taking games. Typically an entire suit is nominated as a trump suit; these cards then outrank all cards of plain (non-trump) suits. The United States of America (USA), commonly known as the United States (U.S.) or America, is a country primarily located in North America. It is a federal union of 50 states and a federal capital district, Washington, D.C.`` \\
        \midrule
        \makecell[tl]{\textbf{LLM-}\\ \textbf{Generated}\\ \textbf{Context}}
        & ``The tweet references Donald Trump's past business practices of seeking H-2B visas to employ foreign workers at his various properties, including Mar-a-Lago. The H-2B visa program allows employers to temporarily hire non-immigrant foreign workers to perform non-agricultural labour or services in the United States. Trump's use of the program has been controversial, given his rhetoric on immigration and American jobs. Critics have argued that he should prioritize hiring American workers instead of seeking foreign labour.'' \\
        \bottomrule
    \end{tabularx}
    \caption{Comparison of entity-linked context vs. LLM-generated context on a sample post.}
    \label{tab:ne-ft-comp}
\end{table}

\clearpage

\begin{figure*}[h]
\vspace{12pt}

  \centering
  
  \begin{subfigure}[b]{0.31\textwidth}
    \centering
    \includegraphics[width=\linewidth, valign=m]{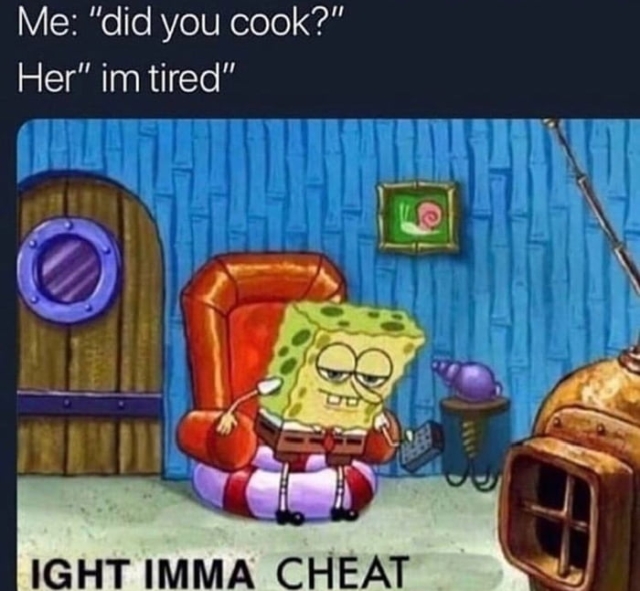}
  \end{subfigure}
  \hfill
  \begin{subfigure}[b]{0.33\textwidth}
    \centering
    \includegraphics[width=\linewidth, valign=m]{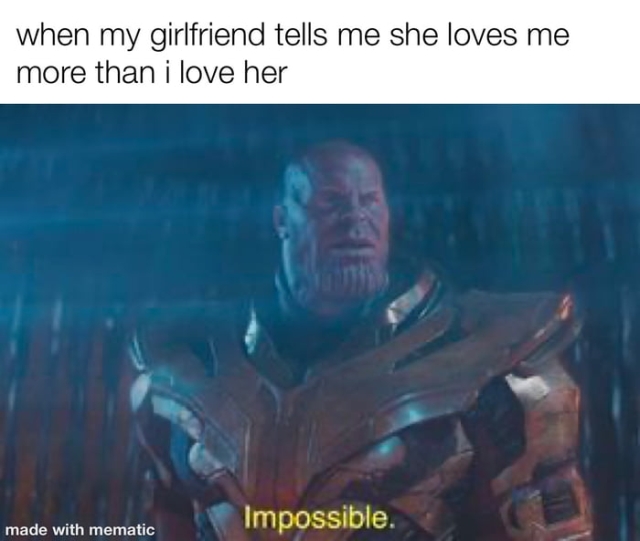}
  \end{subfigure}
  \hfill
  \begin{subfigure}[b]{0.31\textwidth}
    \centering
    \includegraphics[width=\linewidth, valign=m]{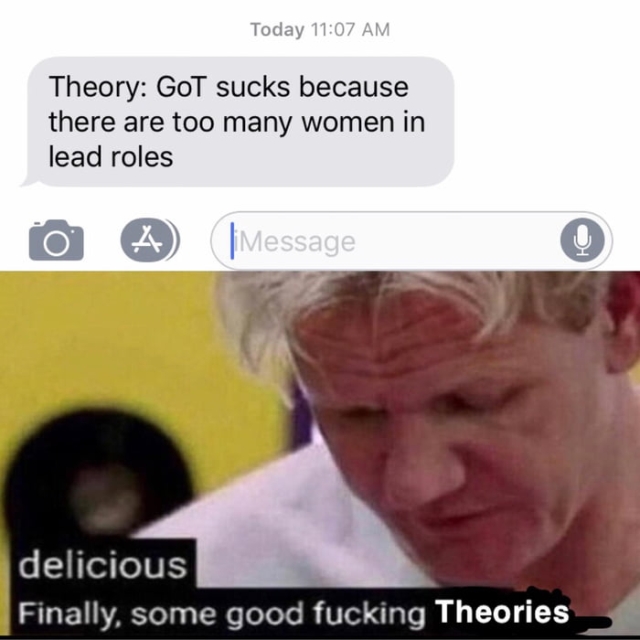}
  \end{subfigure}
  
  \caption{Examples of memes with potentially irrelevant named entities present in the generated image description.}
  \label{fig:irrel-nes}
\end{figure*}

\vspace{24pt}

\begin{table}[h]

    \centering
    \footnotesize
    \begin{tabularx}{\textwidth}{p{3.5cm}X}
        \toprule
        \textbf{Post} & \textbf{Context} \\
        \midrule
        ``berlin is the new jerusalem: more and more israeli jews moving to germany \#swrm via...''
        & The tweet references a perceived increase in Israeli Jews moving to Berlin, Germany. The hashtag ``\#swrm'' likely refers to Südwestrundfunk, a German public broadcaster, suggesting the information may originate from a news report or program by this outlet. Some see Berlin as an attractive destination due to its relatively low cost of living compared to Israel, as well as its vibrant cultural scene and historical significance. \\
        \midrule
        ``\#aryan men \& women will either rise up to fight for their heritage children \& future  or the white race will cease to exist. choose now.''
        & The tweet uses the term ``aryan'', which is associated with white supremacist ideology, and warns that the ``white race will cease to exist'' unless people ``rise up to fight''. This language aligns with the ``great replacement'' conspiracy theory, a white nationalist idea that claims white people are being intentionally replaced by non-white people. Such rhetoric is often used to promote fear and incite violence against minority groups. \\
        \midrule
        \makecell[lt]{``Fight me,\\ communists.''}
        & The tweet ``Fight me, communists'' likely comes from someone with anti-communist views, engaging in political antagonism. Communism is a political and economic ideology advocating a classless society in which the community owns all property and resources. The tweet could be interpreted as a challenge to those who espouse communist beliefs, potentially escalating into online or real-world conflict.\\
        \midrule
        \makecell[lt]{``yes kill all\\ christians''}
        & The statement ``yes kill all christians'' expresses support for violence against Christians, a religious group with billions of adherents worldwide. Christianity has a long and complex history, and Christians have been both perpetrators and victims of violence throughout history. Calls for violence against religious groups are considered hate speech by many definitions.\\
        \midrule
        ``this is what becoming a groyper does to your life people it is an irreconcilable decision''
        & ``Groyper'' is a term used to describe members of the Groypers, a far-right, white nationalist, and alt-right movement in the United States. Groypers are known for their support of white supremacist and anti-Semitic ideologies, as well as their attempts to inject these views into mainstream conservative politics. The tweet likely refers to the consequences or changes in lifestyle and relationships that may occur when someone becomes involved with this movement.\\
        \bottomrule
    \end{tabularx}
    \caption{Post and generated context of Latent Hatred error analysis examples.}
    \label{tab:doubtful}
\end{table}

\clearpage

\begin{table}[h]
\centering

    \begin{tabularx}{\textwidth}{p{4cm} p{3cm} X}
    \toprule
    Meme & Image Description & Context\\
    \midrule
    
    \begin{minipage}[tl]{\linewidth}
        \centering
        \includegraphics[width=\linewidth]{4866.jpg}
    \end{minipage}
    &
    \begin{minipage}[tl]{\linewidth}
        \footnotesize
        ``A cat reaches up to high-five a person's hand.''
    \end{minipage}
    &
    \begin{minipage}[tl]{\linewidth}
        \footnotesize
        ``The meme uses an image of a cat "high-fiving" a person's hand. Text is overlaid on the image to indicate the cat is setting the table and the person is the poster's mom, who is cooking the food. The context is likely intended to be humorous, portraying the traditional division of labour within a household, where the mother cooks and other family members handle simpler tasks. It is a lighthearted observation of domestic roles.''
    \end{minipage}
    \\
    \midrule

    \begin{minipage}[tl]{\linewidth}
        \centering
        \includegraphics[width=\linewidth]{10984.jpg}
    \end{minipage}
    &
    \begin{minipage}[tl]{\linewidth}
        \footnotesize
        ``A crying woman is holding her hand to her forehead.''
    \end{minipage}
    &
    \begin{minipage}[tl]{\linewidth}
        \footnotesize
        ``This meme is using the "Woman Crying" meme format. The text suggests a joke about domestic violence against women. The humor relies on the idea that the woman has been physically abused (given two black eyes) and that further communication would be pointless or repetitive.''
    \end{minipage}
    \\
    \midrule
    
    \begin{minipage}[tl]{\linewidth}
        \centering
        \includegraphics[width=\linewidth]{1322.jpg}
    \end{minipage}
    &
    \begin{minipage}[tl]{\linewidth}
        \footnotesize
        ``A woman in red heels and jean shorts leans over and looks at the exposed engine of a car.''
    \end{minipage}
    &
    \begin{minipage}[tl]{\linewidth}
        \footnotesize
        ``The meme features a woman in a pin-up style outfit looking at a car engine. The text "What am I looking at? This isn't the kitchen" is superimposed on the image. It is a joke based on a sexist stereotype that a woman's place is in the kitchen, and therefore she would not know anything about cars.''
    \end{minipage}
    \\
    \midrule

    \begin{minipage}[tl]{\linewidth}
        \centering
        \includegraphics[width=\linewidth]{images/4970.jpg}
    \end{minipage}
    &
    \begin{minipage}[tl]{\linewidth}
        \footnotesize
        ``A smiling woman holds a mug while a man drinks from one.''
    \end{minipage}
    &
    \begin{minipage}[tl]{\linewidth}
        \footnotesize
        ``The meme features a picture of Carl Winslow and his wife Harriette from the sitcom \textit{Family Matters}. The text suggests a scenario where someone is initially happy or carefree ("laughing") but will face negative consequences or regret after consuming the drink. It is likely used to humorously warn against overindulgence or actions that seem fun at the moment but will lead to a negative outcome.''
    \end{minipage}
    \\
    
    \midrule
    \begin{minipage}[tl]{\linewidth}
        \centering
        \includegraphics[width=\linewidth]{images/8206.jpg}
    \end{minipage}
    &
    \begin{minipage}[tl]{\linewidth}
        \footnotesize
        ``This image shows four illustrations of women, each in a different style and context, conveying the idea that different things empower different women.''
    \end{minipage}
    &
    \begin{minipage}[tl]{\linewidth}
        \footnotesize
        ``The meme is a four-panel image featuring historical artwork. Each panel displays a different image of a woman, accompanied by a caption describing an action or state of being that purportedly empowers women. It is likely designed to express the idea that feminism supports a variety of choices for women, as their empowerment may come from diverse actions, behaviours, and lifestyles.''
    \end{minipage}
    \\

    \bottomrule
    \end{tabularx}
\caption{Generated image descriptions and context of example MAMI memes.}
\vspace{6pt}
\label{multimodal-examples}

\end{table}

\clearpage

\subsection{Additional Results} \label{full-res}

\makeatletter
\renewcommand\subsubsection{\@startsection{subsubsection}{3}{\z@}%
  {-8\p@ \@plus -4\p@ \@minus -4\p@}%
  {4\p@ \@plus 2\p@ \@minus 2\p@}%
  {\normalfont\normalsize\bfseries}}
\makeatother

\vspace{8pt}
\subsubsection*{Latent Hatred Results}
\vspace{24pt}

\begin{table}[H]
\centering
    \begin{threeparttable}

    \footnotesize
    
    \centering
    \begin{tabular}{l
        p{2.8cm}
        >{\centering\arraybackslash}p{0.83cm}
        >{\centering\arraybackslash}p{0.83cm}
        >{\centering\arraybackslash}p{0.83cm}
        >{\centering\arraybackslash}p{0.83cm}
        >{\centering\arraybackslash}p{0.83cm}
        >{\centering\arraybackslash}p{0.83cm}}
        \toprule
        \multirow{2}{*}{\textbf{Context}} 
        & \multirow{2}{*}{\makecell[l]{\textbf{Incorporation}\\ \textbf{Strategy}}} 
        & \multicolumn{3}{c}{\textbf{Binary}} 
        & \multicolumn{3}{c}{\textbf{Multi-Class}} \\
        & & \textbf{P} & \textbf{R} & \textbf{F1} & \textbf{P} & \textbf{R} & \textbf{F1} \\
        \toprule
        Zero-Context & - & 0.73 & 0.72 & 0.72 & \textbf{0.51} & \textbf{0.51} & \textbf{0.51} \\
        REL \cite{linLeveragingWorldKnowledge2022} & Append \& Embed & 0.71 & 0.70 & 0.70 & 0.48 & 0.47 & 0.47 \\
        ConceptNet \cite{elsheriefLatentHatred2021} & Embed \& Concat & \textbf{0.74} & \textbf{0.74} & \textbf{0.74} & 0.50 & 0.50 & 0.50 \\
        LLM Prediction & - & 0.74 & 0.70 & 0.70 & 0.40 & 0.28 & 0.26 \\
        \midrule
        \multirow{4}{*}{LLM-Generated} 
        & Append \& Embed & 0.73 & 0.72 & 0.73 & \textbf{0.54} & \textbf{0.53} & \textbf{0.53} \\
        & Embed \& Concat & \textbf{0.75} & \textbf{0.75} & \textbf{0.75} & 0.52 & 0.52 & 0.52 \\
        & Context-Embed & 0.71 & 0.71 & 0.71 & 0.47 & 0.47 & 0.47 \\
        & LLM Enhance & 0.71 & 0.71 & 0.71 & 0.49 & 0.49 & 0.49 \\
        \bottomrule
    \end{tabular}
    
    \caption{Macro-averaged Precision, Recall, and F1 scores for binary and implicit HSD on the Latent Hatred dataset.}
    \end{threeparttable}
\end{table}

\vspace{18pt}

\begin{table}[H]
\centering
    \begin{threeparttable}
    
    \footnotesize
    
    \centering
    \begin{tabular}{llcccccc}
        \toprule
        \textbf{Context} & \makecell[l]{\textbf{Incorporation}\\\textbf{Strategy}} 
        & \makecell[c]{\textbf{Incite-}\\\textbf{ment}} 
        & \makecell[c]{\textbf{Infer-}\\\textbf{iority}} 
        & \textbf{Irony}
        & \makecell[c]{\textbf{Stereo-}\\\textbf{typing}} 
        & \makecell[c]{\textbf{Threat-}\\\textbf{ening}}
        & \makecell[c]{\textbf{White}\\\textbf{Grievance}} \\
        \toprule
        Zero-Context & - & \textbf{0.53} & 0.51 & \textbf{0.56} & \textbf{0.57} & \textbf{0.57} & \textbf{0.61} \\
        REL \cite{linLeveragingWorldKnowledge2022} & Append \& Embed & 0.51 & 0.47 & 0.53 & 0.52 & 0.55 & 0.57 \\
        ConceptNet \cite{elsheriefLatentHatred2021} & Embed \& Concat & \textbf{0.53} & \textbf{0.54} & \textbf{0.56} & 0.56 & 0.56 & 0.60 \\
        LLM Prediction & - & 0.08 & 0.40 & 0.24 & 0.39 & 0.11 & 0.60 \\
        \midrule
        \multirow{4}{*}{LLM-Generated} 
        & Append \& Embed & \textbf{0.55} & \textbf{0.53} & 0.56 & \textbf{0.61} & 0.59 & \textbf{0.63}  \\
        & Embed \& Concat & \textbf{0.55} & \textbf{0.53} & \textbf{0.58} & 0.57 & \textbf{0.61} & 0.61 \\
        & Context-Embed & 0.52 & 0.47 & 0.49 & 0.51 & 0.50 & 0.60  \\
        & LLM Enhance & 0.51 & 0.50 & 0.53 & 0.55 & 0.56 & 0.60  \\
        \bottomrule
    \end{tabular}
    \caption{Per-class F1 of implicit multi-class HSD on the Latent Hatred dataset.}
    \end{threeparttable}
\end{table}

\clearpage

\subsubsection*{MAMI Results}
\vspace{24pt}

\begin{table}[H]
\centering
    \begin{threeparttable}
    \footnotesize
    \centering
    \begin{tabular}{l
        p{2.8cm}
        >{\centering\arraybackslash}p{0.8cm}
        >{\centering\arraybackslash}p{0.8cm}
        >{\centering\arraybackslash}p{0.8cm}
        >{\centering\arraybackslash}p{0.8cm}
        >{\centering\arraybackslash}p{0.8cm}
        >{\centering\arraybackslash}p{0.8cm}}
        \toprule
        \multirow{2}{*}{\textbf{Context}} 
        & \multirow{2}{*}{\makecell[l]{\textbf{Incorporation}\\ \textbf{Strategy}}} 
        & \multicolumn{3}{c}{\textbf{Binary}} 
        & \multicolumn{3}{c}{\textbf{Multi-Label}} \\
        & & \textbf{P} & \textbf{R} & \textbf{F1} & \textbf{P} & \textbf{R} & \textbf{F1} \\
        \toprule
        Zero-Context & - & 0.79 & 0.79 & 0.79 & 0.59 & 0.58 & 0.59 \\
        REL \cite{linLeveragingWorldKnowledge2022} & Append \& Embed & 0.78 & 0.78 & 0.78 & 0.58 & 0.56 & 0.57 \\
        ConceptNet \cite{elsheriefLatentHatred2021} & Embed \& Concat & 0.80 & 0.80 & 0.80 & \textbf{0.60} & 0.60 & \textbf{0.60} \\
        LLM Prediction & - & \textbf{0.86} & \textbf{0.86} & \textbf{0.86} & 0.56 & \textbf{0.68} & \textbf{0.60} \\
        \midrule
        \multirow{4}{*}{LLM-Generated} 
        & Append \& Embed & 0.84 & 0.84 & 0.84 & 0.61 & 0.62 & 0.62 \\
        & Embed \& Concat & \textbf{0.85} & \textbf{0.85} & \textbf{0.85} & 0.62 & \textbf{0.63} & \textbf{0.63} \\
        & Context-Embed & 0.83 & 0.83 & 0.83 & 0.62 & 0.60 & 0.61 \\
        & LLM Enhance & 0.83 & 0.83 & 0.83 & \textbf{0.64} & 0.61 & 0.62 \\
        \bottomrule
    \end{tabular}
    \caption{Macro-averaged Precision, Recall, and F1 scores for binary and multi-label HSD on the MAMI dataset.}
    \label{mami-extra-results}
    \end{threeparttable}
\end{table}

\vspace{18pt}

\begin{table}[H]
\centering
    \begin{threeparttable}
    \footnotesize  % Reduce font size
    \centering
    \begin{tabular}{llccccc}
        \toprule
        \textbf{Context} 
        & \makecell[l]{\textbf{Incorporation}\\\textbf{Strategy}} 
        & \textbf{Shaming}
        & \makecell[c]{\textbf{Stereo-}\\\textbf{typing}}
        & \makecell[c]{\textbf{Object-}\\\textbf{ification}}
        & \textbf{Violence} \\
        \toprule
        Zero-Context & - & 0.45 & 0.68 & 0.67 & 0.54 \\
        REL \cite{linLeveragingWorldKnowledge2022} & Append \& Embed & 0.42 & 0.68 & 0.65 & 0.53 \\
        ConceptNet \cite{elsheriefLatentHatred2021} & Embed \& Concat & 0.43 & \textbf{0.69} & 0.68 & 0.59 \\
        LLM Prediction & - & \textbf{0.46} & 0.60 & \textbf{0.71} & \textbf{0.63}\\
        \midrule
        \multirow{4}{*}{LLM-Generated} 
        & Append \& Embed & \textbf{0.49} & 0.69 & 0.69 & 0.58 \\
        & Embed \& Concat & \textbf{0.49} & 0.68 & \textbf{0.71} & \textbf{0.65} \\
        & Context-Embed & 0.45 & 0.68 & 0.70 & 0.58 \\
        & LLM Enhance & 0.46 & \textbf{0.71} & \textbf{0.71} & 0.59 \\
        \bottomrule
    \end{tabular}
    \caption{Per-label F1 for multi-label classification on the MAMI dataset.}
    \end{threeparttable}
\end{table}

\clearpage
\subsection{Confusion Matrices} \label{conf-mat}

\vspace{8pt}
\subsubsection*{Latent Hatred Confusion Matrices}
\vspace{12pt}

\begin{figure}[h!]
  \centering
  \begin{subfigure}[b]{0.48\textwidth}
    \centering
    \includegraphics[width=0.9\textwidth]{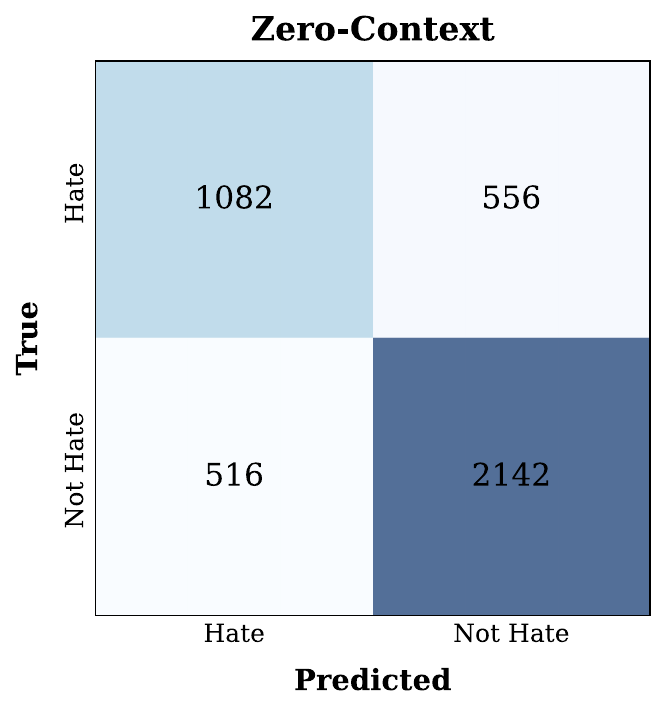}
  \end{subfigure}
  \hfill
  \begin{subfigure}[b]{0.48\textwidth}
    \centering
    \includegraphics[width=0.9\textwidth]{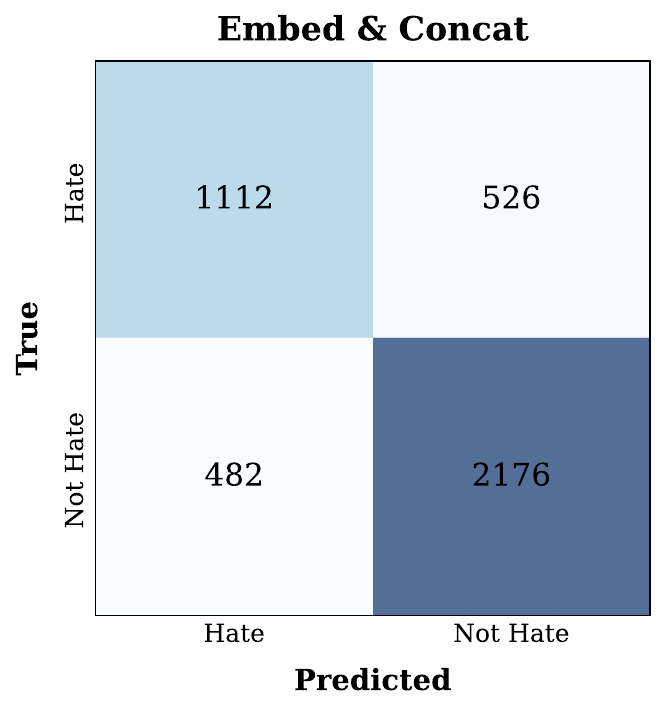}
  \end{subfigure}
  \caption{Confusion matrices for Zero-Context (left) and LLM-Generated Embed \& Concat (right) binary HSD predictions on the Latent Hatred dataset.}
\end{figure}

\vspace{12pt}

\begin{figure}[h!]
  \centering
  \begin{subfigure}[b]{0.48\textwidth}
    \centering
    \includegraphics[width=\textwidth]{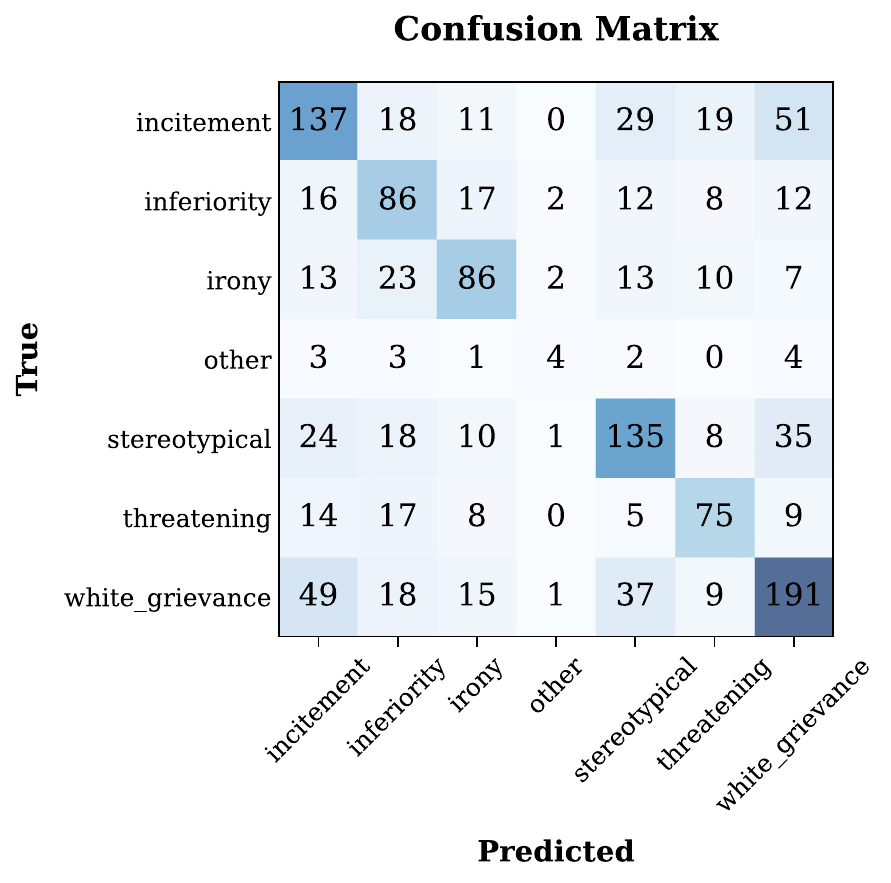}
  \end{subfigure}
  \hfill
  \begin{subfigure}[b]{0.48\textwidth}
    \centering
    \includegraphics[width=\textwidth]{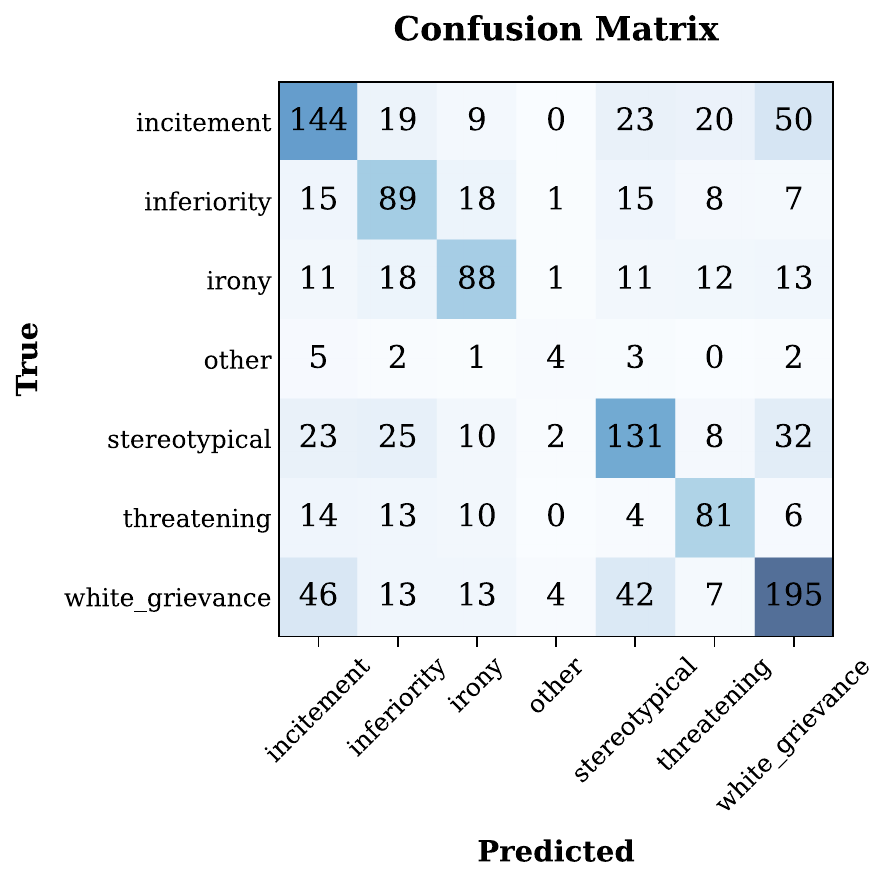}
  \end{subfigure}
  \caption{Confusion matrices for Zero-Context (left) and LLM-Generated Embed \& Concat (right) implicit multi-class HSD predictions on the Latent Hatred dataset.}
\end{figure}

\clearpage

\vspace{8pt}
\subsubsection*{MAMI Confusion Matrices}
\vspace{12pt}

\begin{figure}[h!]
  \centering
  \begin{subfigure}[b]{0.48\textwidth}
    \centering
    \includegraphics[width=0.9\textwidth]{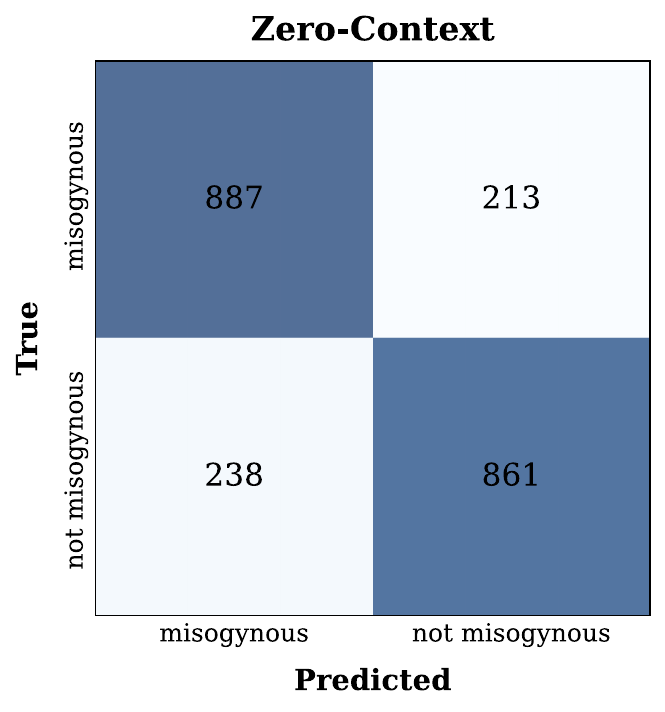}
  \end{subfigure}
  \hfill
  \begin{subfigure}[b]{0.48\textwidth}
    \centering
    \includegraphics[width=0.9\textwidth]{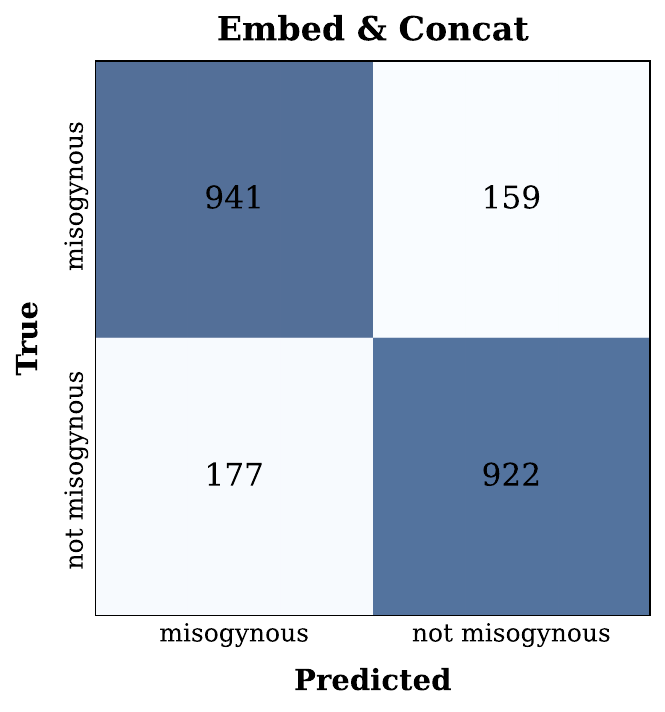}
  \end{subfigure}
  \caption{Confusion matrices for Zero-Context (left) and LLM-Generated Embed \& Concat (right) binary HSD predictions on the MAMI dataset.}
\end{figure}

\vspace{12pt}

\begin{figure}[h!]
  \centering
  \includegraphics[width=\textwidth]{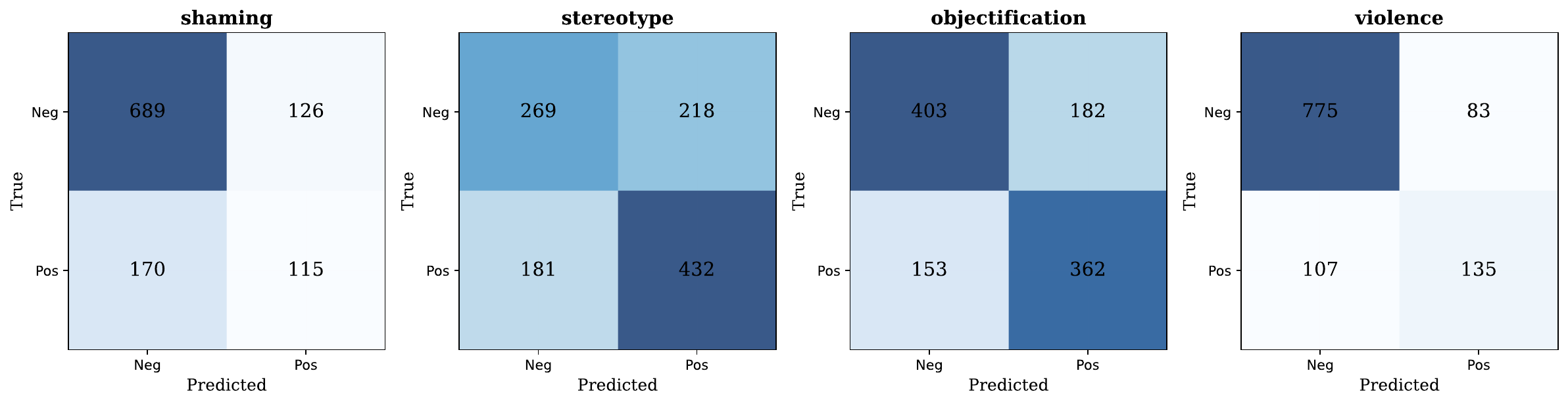}
  \caption{Confusion matrices for Zero-Context multi-label predictions on the MAMI dataset.}
  \vspace{14pt}
  \includegraphics[width=\textwidth]{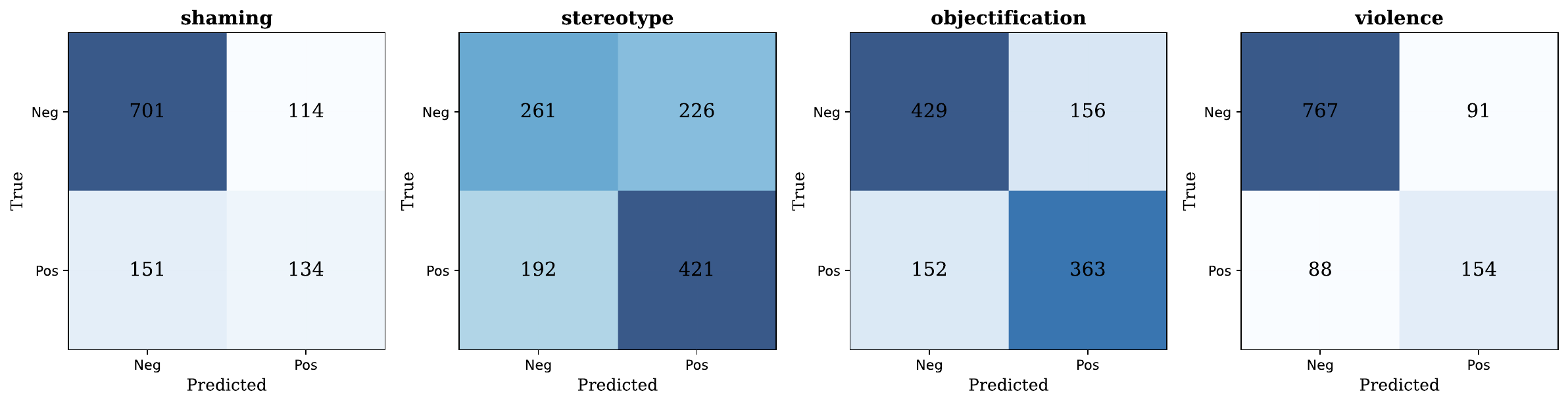}
  \caption{Confusion matrices for LLM-Generated Embed \& Concat multi-label predictions on the MAMI dataset.}
  \label{imp-dist}
\end{figure}

\end{document}